# Frequency, acceptability, and selection: A case study of clause-embedding

Aaron Steven White
*University of Rochester*
aaron.white@rochester.edu

Kyle Rawlins
*Johns Hopkins University*
kgr@jhu.edu

**Abstract** We investigate the relationship between the frequency with which verbs are found in particular subcategorization frames and the acceptability of those verbs in those frames, focusing in particular on subordinate clause-taking verbs, such as *think*, *want*, and *tell*. We show that verbs' subcategorization frame frequency distributions are poor predictors of their acceptability in those frames—explaining, at best, less than $\frac{1}{3}$ of the total information about acceptability across the lexicon—and, further, that common matrix factorization techniques used to model the acquisition of verbs' acceptability in subcategorization frames fare only marginally better.

## 1 Introduction

Knowing a language involves at least being able to judge the well-formedness of strings relative to that language. Accounts differ on a variety of axes: (i) whether or not there is a psychologically real distinction between notions of well-formedness, such as *grammaticality* and *acceptability* (Chomsky 1965 *et seq*); (ii) whether either (or both) concepts are discrete or continuous; and (iii) what exactly is required to collect the supporting judgments (Bard et al. 1996; Keller 2000; Sorace & Keller 2005; Sprouse 2007; 2011; Featherston 2005; 2007; Gibson & Fedorenko 2010; 2013; Sprouse & Almeida 2013; Sprouse et al. 2013; Schütze & Sprouse 2014; Lau et al. 2017; Sprouse et al. 2018 among many others). But few doubt that some specification of well-formedness is a crucial component of a theory of linguistic knowledge.

Well-formedness is something that must be learned—at least in part—from linguistic experience. A major question is how direct the relationship between well-formedness and linguistic experience is. Approaches that directly address this question have tended to focus on complex syntactic effects—e.g. effects that might arise from constraints on syntactic movement, such as island effects (see Kluender & Kutas 1993; Hofmeister & Sag 2010; Sprouse et al. 2012; Kush et al. 2018 among others). The abstractness of these constraints makes them useful test cases, since they are prime candidates for knowledge that might arise from inductive biases innate to language learners—i.e. not directly from statistical properties of the input (though see Pearl & Sprouse 2013). However, complex syntactic phenomena are far from the only factor contributing to well-formedness.

Effects on well-formedness that are at least partially a product of lexical knowledge have garnered much less focused attention within theoretical linguistics—possibly, because they seem more directly dependent on statistical properties of the input (though see Bresnan 2007; Bresnan et al. 2007; White 2015; White et al. 2017a).[1] That is, lexical, as opposed to grammatical, constraints on well-formedness might have a direct connection to co-occurrence statistics in language learners' input, and therefore be learnable using relatively simple strategies—e.g. tracking co-occurrence frequencies, which has long been believed to be well within the capabilities of even young children (Saffran et al. 1996b; a; Aslin et al. 1998; Maye et al. 2002). As such, lexical constraints on well-formedness could in principle make a better case for a direct connection between linguistic experience and grammatical knowledge.

Against this background, this paper makes two main contributions: we introduce and validate an empirical method for collecting lexical acceptability patterns at a very large scale, and we use a dataset collected via this method to computationally investigate the connection between linguistic experience (in the form of corpus data) and linguistic knowledge (in the form of acceptability patterns). The main finding from this second part is that knowledge of acceptability does *not* provide support for a direct connection between frequency and acceptability, but rather supports the idea that a language learner needs to employ substantial abstraction in order to be able to achieve adult/human-like knowledge of acceptability.

[1] See also much work in the sentence processing literature (Trueswell et al. 1993; Spivey-Knowlton & Sedivy 1995; Garnsey et al. 1997; McRae et al. 1998; Altmann & Kamide 1999; Hale 2001; Levy 2008; Wells et al. 2009; Fine & Jaeger 2013; Linzen & Jaeger 2016 among others) and the language acquisition literature (Landau & Gleitman 1985; Pinker 1984; 1989; Gleitman 1990; Naigles 1990; 1996; Naigles et al. 1993; Fisher et al. 1991; Fisher 1994; Fisher et al. 1994; 2010; Lederer et al. 1995; Gillette et al. 1999; Snedeker & Gleitman 2004; Lidz et al. 2004; Gleitman et al. 2005; Papafragou et al. 2007 among others).

In particular, we investigate effects on well-formedness having to do with verbs' c(ategory)-selection behavior—i.e. what kinds of syntactic structures verbs are acceptable in—showing that this assumption is not justified. We focus specifically on verbs that take subordinate clauses—henceforth, *clause-embedding verbs*—such as *think*, *want*, and *tell*.

(1) a. Jo {thought, told Mo} that Bo left.
b. Jo {wanted, told} Bo to leave.

Clause-embedding verbs are a useful test case because (i) subordinate clauses can have a wide variety of syntactic structures; (ii) many verbs can take a large subset of these clause types; and (iii) there is high variability in which subset of clause types verbs can take. For instance, *remember* can combine with a wide variety of differently structured clauses, as in (2a)-(2e) as well as noun phrases (2f) and the intransitive frame in (2g).

(2) a. Jo remembered that Bo left.
b. Jo remembered Bo to have left.
c. Jo remembered Bo leaving.
d. Jo remembered to leave.
e. Jo remembered leaving.
f. Jo remembered Bo.
g. Jo remembered.

Starting from the assumption that the relevant lexical knowledge must be acquired, in some way or other, from the co-occurrence statistics, our investigation specifically aims to measure how direct a relationship there is between verbs' selectional patterns—as measured by *acceptability judgments*—and said statistics. We take this question to be one about the lexicon as a whole, and therefore test it with a *lexicon-scale* dataset of acceptability judgments for clause-embedding verbs. A lexicon-scale investigation is both possible and crucial: there are many clause-embedding verbs in English—by some counts, at least 1,000 (White & Rawlins 2016)—so testing grammatical knowledge involves testing knowledge across this large set.

Our investigation contrasts with prior work, which investigates only a small set of key verbs and frames (Fisher et al. 1991; Lederer et al. 1995; Bresnan et al. 2007; White et al. 2018a; though see Kann et al. 2019). The relatively small size of previous investigations is likely a product of the fact that scaling standard methodologies to a larger set of verbs is infeasible without introducing unwanted biases or insurmountable workload. To overcome this obstacle, we propose a novel method—the *bleaching method*— for automatically scaling standard methods while avoiding the introduction of such biases into the item set.

In Section 3, we report on an experiment validating the bleaching method against a more standard acceptability judgment collection method, focusing on a small set of clause-embedding verbs. In Section 4, we report on an experiment in which we deploy the bleaching method on 1,000 clause-embedding verbs in 50 syntactic frames to create the MegaAcceptability dataset, which was first reported on in White & Rawlins 2016 and is publicly available at megaattitude.io under the auspices of the MegaAttitude Project.[2] In Section 5, we use the MegaAcceptability dataset in conjunction with a very large dataset of verbs' subcategorization frequencies to show that the relationship between acceptability and frequency, when considering this entire sublexicon, is surprisingly weak. This throws into question the assumption that c-selectional behavior can be directly read off frequency distributions.

Nonetheless, adult native speakers are able to judge the acceptability of the items that make up this task. If they are unable to learn the information to do this directly from the frequency—even assuming access to the data that is both ideal and uniform—they must have *some* way to get it. This suggests that some abstraction of the frequency distributions in the input is necessary. In Section 6, we consider a variety of such abstractions, showing (i) that common, shallow factorization methods yield miniscule improvements in the prediction of acceptability over more direct models; but (ii) that methods involving multiple layers of abstraction can predict acceptability quite well. The models we present here are proof-of-concept models rather than attempts at actual learning models, and thus, our results do not fully answer the question of exactly how a learner will gather this information; but we take them to strongly confirm the necessity of substantial abstraction in grammatical theory relative to input frequency. In Section 7, we conclude with remarks on what these findings imply for the acquisition of distributional knowledge about lexical items.

[2] As discussed in Section 4, this dataset has appeared in brief form in three proceedings papers: White & Rawlins 2016, where it was introduced in the context of building a computational model to infer semantic types, and White & Rawlins (2018); White et al. (2018b), where it was used as a starting point for developing a data set about veridicality that is not relevant here. The present paper reviews in much greater detail the methods for constructing the data set, and presents validation experiments and arguments in favor of the bleaching method that have not previously appeared.

# 2 Background

We begin with a discussion of previous work relating frequency and acceptability (Section 2.1), acceptability and selection (Section 2.2), and frequency and selection (Section 2.3) and then discuss two hypotheses about the joint relationship among the three (Section 2.4).

## 2.1 *Frequency and acceptability*

In a substantial body of work, Clark, Lappin, and colleagues argue that knowledge of grammaticality, by way of acceptability, can be modeled, to a large extent, with probabilistic models that involve a direct link between probability and acceptability and directly recognize gradience in acceptability (Clark & Lappin 2011; Clark et al. 2013a; b; Lau et al. 2017; see also Bresnan et al. 2007; Bresnan 2007). These models have two components: (i) a way of estimating the probability of a sentence of a language; and (ii) a way of translating probabilities to acceptabilities. For the most part, the latter is quite straightforward: some variant of log probabilities, normalized for sentence length and unigram frequency effects. The former is where much of the action is.

This body of work considers a range of possibilities for how to model probability, ranging from simple *n*-gram frequency models to neural language models in Lau et al. 2017 (see also Warstadt et al. 2019). Lau et al. consider two kinds of data: (i) sentences sampled from the BNC corpus, fed through google translate to a range of languages, and then translated back to English (with the goal of obtaining a spread of acceptability), and a sample of sentences from Adger 2003 by way of Sprouse & Almeida's (2013) dataset. They then run a variety of acceptability judgment studies on Mechanical Turk, providing the core data for their experiments. Across the board, the models in this paper show what the authors describe as an "encouraging degree of accuracy" in predicting human judgments of acceptability—well exceeding baseline models but generally falling well short of human performance.

The authors take this to be a signal that probabilistic models are the current best way to incorporate gradience in human judgments about acceptability into a theory of grammatical knowledge. This conclusion is controversial (Sprouse et al. 2018), but we take it as a starting point that current probabilistic models can at least do reasonably well at capturing many—though probably not all—facets of acceptability.

Clark, Lappin, and colleagues' work therefore sets the stage for the central question that we are addressing here: what is the relationship between frequency of use, probability, and acceptability? Though the model generating the probabilities itself may be complicated, on Lau et al.'s view, the relationship between the two is *direct*. That is, there is a simple transformation that, as long as some basic normalization is taken care of, more or less directly predicts acceptability.

Existing work has focused on testing hypotheses like this on data that is extremely varied—e.g. random samples of naturalistic corpus data or broad datasets of grammaticality judgments from linguistic theory (Sprouse et al. 2018). For this reason, it is an independently challenging and interesting problem to estimate the probability of sentences across such data, and as Lau et al. (2017) demonstrate, in at least some domains more sophisticated models from natural language processing (NLP) will lead to better predictions of acceptability—e.g. models which involve more complex relationships between the frequencies they are trained on and the probabilities they output. But this approach to the underlying data leads to an additional problem: conclusions about the probabilistic nature of the grammar rest on the degree to which the $[0, 1]$ interval values that these models are producing are in fact good estimates of the probabilities of particular sentences, also making it rather challenging to know what the driving factor for sentence-level probabilities is.

In the present work, we take a different approach. Rather than a broad and diverse data sample, we pick a single phenomenon where we can obtain acceptability data exhaustively, and estimate probabilities from corpora in a relatively transparent way, producing a direct idealization of linguistic experience.[3] The particular data we investigate is selectional patterns for clause-embedding predicates, where the main point of variation between items is just the verb and its selectional frame. This also allows us to begin to localize the *kinds* of grammatical knowledge that are likely to be involved in variation in acceptability.

[3] This is analogous to recent approaches within the NLP literature that aim to probe what linguistic knowledge different models learn from corpus data using an array of focused datasets (Linzen et al. 2016; White et al. 2017b; 2018b; Gulordava et al. 2018; Kuncoro et al. 2018; Peters et al. 2018; Poliak et al. 2018; Conneau et al. 2018; Wang et al. 2018; Wilcox et al. 2018; McCoy et al. 2019; Kann et al. 2019: a.o.).

## 2.2 Acceptability and selection

The selectional behavior of verbs in general—and clause-embedding verbs in particular—is a classic topic in linguistic theory (Chomsky 1965; Gruber 1965; Fillmore 1970; Jackendoff 1972; Chomsky 1973; Grimshaw 1979; 1990; Pesetsky 1982; 1991 among many others). There are broadly accepted to be two crucial descriptive factors leading to variance in whether a verb is acceptable in a particular sentence (other things being equal): (i) semantic constraints imposed by the argument structure of the verb (*s(emantic)-selection*); and (ii) (morpho-)syntactic constraints imposed by a verb on its complements (*c(omplement)-selection*; Grimshaw 1979).

It is a matter of substantial debate and discussion whether some or all of these constraints might be predictable from each other or other factors—e.g. event structure—something we will not try to settle here (see Levin & Rappaport Hovav 2005 and references therein). But this domain is useful for our purposes because (i) it is empirically rich—even if we focus just on selection of clauses; and (ii) it involves both a large set of patterns and a large set of lexical idiosyncracies.

For example, Grimshaw introduced the classic *wonder* v. *think* comparison (her Ex. 1):

(3) a. John wondered who Bill saw.
b. *John wondered that Bill saw someone.
c. John thought that Bill saw someone.
d. *John thought who Bill saw.

We suggest that this kind of data, when scaled up to the entire lexicon, is a perfect test-bed for questions about acceptability and grammar. In this case, the two verbs differ inversely in whether they license interrogative vs. declarative complements (though see White 2019), something we expect would be mirrored in corpus frequencies (after controlling for, e.g., the fact that *think* is itself much more frequent than *wonder*).

While there are many more frames than just these two, there are a limited number of possible selectional patterns instantiated in English; and while, as we have suggested, there are *many* more verbs that might potentially participate in patterns like this, this number (about 1,000) is still tractable with modern experimental and corpus methods. Further, there is a large body of literature arguing that such patterns involve many subregularities that a learner could find, and the key points of between-sentence variation involved in pairs like this are fairly minimal and (relatively) pos-sible to extract from corpus data. Finally, we would expect some gradience in the judgments—e.g. (3b) is sometimes claimed to be 'not as bad' as (3d).

To date, there is not a large body of experimental linguistics work on acceptability and selection. But the role of selectional patterns has been crucial in research on language acquisition—and particularly verb learning since Landau & Gleitman's (1985) and Gleitman's (1990) seminal work. This line of work suggests that children use syntactic frame information to infer semantic representation when learning the meanings of verbs. We do not review this literature in detail (see White 2015 for a recent review), but since the acceptability judgment method in Fisher et al. 1991 (experiment 1, part B) is a crucial predecessor to what we develop here, we briefly discuss it (see also Lederer et al. 1995).

In Fisher et al.'s (1991) method, a set of verbs and a set of syntactic frames are selected, and the full cartesian product of these sets is constructed. For each pair in the cartesian product one or more sentences are produced by instantiating the syntactic frames with lexical items, then placing the past tense form of the verb in the resulting instantiation—unless the frame is explicitly specified for some other tense/modal. Subjects then do an acceptability judgment task that (across all subjects) exhausts the matrix of verb-frame pairs.

In this study, Fisher et al. used carefully hand-constructed non-idiomatic sentences for each cell of this matrix: 24 verbs $\times$ 39 frames = 936 sentences. For many purposes, this method can work well as long as the verbs and frames are appropriately sampled relative to the questions at hand; but compared to the total scale of the lexicon, it does not even come close to being exhaustive. Moreover, the method does not easily scale much beyond the size of the original Fisher et al. study because of the challenge of hand-constructing items; for 1000 verbs this would require construction 39,000 sentences. While it may be possible to automate this process to some degree, as Fisher et al. (1991) note (p. 347), substantial care and effort is required in selecting the sentences to be judged because of the potential for unintended item effects. Thus, the challenge is to understand how it might be possible to generate sentences on this scale without requiring hand-inspection of every such sentence. We take this challenge on in Sections 3 and 4.

## 2.3 Frequency and selection

The connection between frequency and acceptability in the domain of clause selection is not in general well-studied—or really, even on the radar of the theoretical linguistics work mentioned above (though see Bresnan 2007; Bresnan et al. 2007 and references in fn. 1). But as for acceptability and selection, frequency of exposure has played a role in discussions of the acquisition of verb meaning. In particular, it has been hypothesized that the frequency with which a verb occurs in a syntactic structure (along with the frequency of that syntactic structure across verbs) plays a role in learning that verb's meaning (Lederer et al. 1995; Alishahi & Stevenson 2008; Barak et al. 2012; White 2015; White et al. 2017a).

An important component of these proposals are the mechanisms they employ for normalizing frequency information across verbs—in particular, the cooccurrence frequencies for verbs in different syntactic structures—and subsequently abstracting that frequency information. These mechanisms tend to take the form of clustering models (Lederer et al. 1995; Schulte im Walde 2006), mixture models (Alishahi & Stevenson 2008; Barak et al. 2012), or matrix factorization models (White 2015; White et al. 2017a). We defer detailed discussion of these models until Sections 5 and 6.

## 2.4 Hypotheses

Based on the prior work discussed in this section, we suggest the following hypotheses:

**H1** Verb-frame co-occurrence frequencies predict acceptability.

**H2** Verb-frame co-occurrence frequencies require syntactic/semantic abstractions in order to predict acceptability.

Given the findings of this prior work, it would be quite surprising if H1 turns out to be entirely false: there is likely to be some sense in which these frequencies do predict acceptability. However, the linguistics and acquisition literature on verb representations suggests H2—some kind of abstraction may be necessary to learn even something like acceptability patterns, where in principle there is the potential to just match frequencies. For the human learner, the situation is made starker by something we are idealizing away from: a child does not receive the kind of massive, balanced corpus input our models will.

The program for testing these hypotheses is, at this point, straightforward: we need a (large) selectional acceptability dataset, an estimation of frequencies, and models that compare the two by varying the amount of abstraction. We now turn to the first of these.

# 3 The Bleaching Method

A major obstacle to scaling standard acceptability judgment tasks to entire subregions of the lexicon is ensuring that plausibility effects—i.e. effects on acceptability that are driven by how prototypical the situation described by the sentence is, as opposed to effects driven by syntactic well-formedness—are controlled for. We propose a method to control for these effects by 'semantically bleaching' all lexical category words besides a word of interest—in our case, the clause-embedding verb. Specifically, we manipulate the syntactic context that a word appears in while instantiating all NPs in that context with indefinites (*someone*, *something*) and all verbs in that context (besides the one of interest) with a low content eventive (*happen*, *do*) or stative (*have*) verb.

We first demonstrate the validity of this method on a small set of verbs by showing that agreement is high among naïve participants' acceptability ratings when responding to 'contentful sentences' v. when responding to 'bleached sentences' that are otherwise matched in terms of structure—effectively comparing our bleaching method against the Fisher et al.'s (1991) more standard method, described above. We use the data reported in White et al. 2018a as a dataset of acceptability judgments to contentful sentences—as it focuses on exactly the phenomena we are interested in—and collect acceptability judgments to bleached sentences ourselves.

## 3.1 *Materials*

We follow White et al. (2018a) in using the 30 propositional attitude verbs found in (4), which were selected in such a way that they evenly span the verb classes presented in Hacquard & Wellwood 2012.

(4)
a. think
b. realize
c. understand
d. suppose
e. guess
f. expect
g. imagine
h. remember
i. forget
j. see
k. hear
l. feel
m. tell
n. say
o. promise
p. hope
q. worry
r. doubt
s. pretend
t. deny
u. forbid
v. allow
w. promise
x. love
y. hate
z. bother
aa. amaze
bb. demand
cc. want
dd. need

We also follow White et al. in testing frames built from the same combinations of syntactic features—including various tense-aspect combinations within matrix and embedded clauses as well as various forms of NP and PP arguments.[4] They construct 30 subcategorization frames from these combinations of features.[5] These frames are given in abstracted form in (5), with our instantiation for each constituent type in (7). Importantly, these frames cover a wide range of syntactic contexts and do not just limit themselves to frames with clauses in them. This choice is driven by our main research questions, which are about verb knowledge, and so we cannot exclude, e.g., intransitive or simple transitive NP-taking frames for verbs that also take clauses in some cases.

(5)
a. NP ___ed
b. NP ___ed NP
c. NP ___ed NP NP
d. NP ___ed about NP
e. NP ___ed NP about NP
f. NP ___ed so
g. NP ___ed to
h. NP ___ed S
i. NP ___ed that S
j. NP ___ed if S
k. NP ___ed $S_{wh}$
l. NP ___ed NP S
m. NP ___ed NP that S
n. NP was ___ed that S
o. NP ___ed it that S
p. NP ___ed to NP that S
q. NP ___ed for NP to VP
r. NP ___ed to VP
s. NP ___ed WH to VP
t. NP ___ed NP to VP
u. NP was ___ed to VP
v. NP ___ed there to VP
w. NP ___ed $VP_{ing}$
x. NP ___ed NP $VP_{ing}$
y. NP ___ed NP VP
z. It ___ed NP that S
aa. It ___ed NP $S_{wh}$
bb. It ___ed NP WH to VP
cc. It ___ed NP to VP
dd. S, NP ___ed

[4] An anonymous reviewer asks why White et al. 2018a, and by extension us, do not investigate just frames that take clauses. The vast majority of verbs that do take clauses participate in frames where there are no clauses present, and it is not plausible to assume that cases like this involve different verb senses (at least *a priori*). For example, one widely discussed example is that of verbs that take so-called *concealed question* NPs and the relationship of the distribution of those NP frames to full clause-embedding frames. (Heim 1979; Romero 2005; Nathan 2006; Frana 2010 a.o.). Therefore, since the research questions are about lexical representation, not clauses *per se*, we include all syntactic frames that we believe may bear on this lexical representation, not just ones with clauses.

[5] White et al. note that their experiment in fact included some syntactic frames that involve degree modification, though they do not list these frames. White et al. 2014, which presents a preliminary analysis of the dataset presented in White et al. 2018a, lists four such frames (see the Appendix of that paper). We do not include these in our experiment, since the analyses in White et al. 2018a do not include them, and thus our statistics would not be comparable to theirs if we did.

ee. S, I ___ ff. S, NP ___ed

To instantiate these abstract frames, White et al. instantiate each of the phrases with contentful lexical items (following Fisher et al. 1991).[6] For instance, (6) shows one of three items that instantiate the pair (*think*, *NP _ed that S*) in White et al.'s experiment.

(6) Gary thought that she fit the part.

As noted above, one potential issue that arises when using contentful lexical items to instantiate a frame is that ratings of the resulting items are susceptible to typicality effects—some verbs might sound less plausible in certain frame instantiations even if that verb is otherwise perfectly fine in some other instantiation of that same frame. White et al. control for such typical effects by creating three different instantiations for each frame and taking into account possible item variability in their analysis. But if at all possible, it is ideal not to do this, since it increases the number of items and lowers the statistical power of subsequent analyses—thus requiring more ratings to get an accurate estimate of the acceptability of any particular verb-frame pair.

To address this both in the current experiment and in the large-scale experiment reported in Section 4, we instead create only one instantiation of each frame with as little lexical content as possible. All instantiations we use are listed in (7).

(7)
a. **NP** Noun phrase (*someone* or *something*)
b. **VP** Verb phrase with verb in bare form (*do something*)
c. **VP**$_{ing}$ Verb phrase with verb in present progressive form (*doing something*)
d. **S** Full clause without complementizer (*something happened*)
e. **S**$_{wh}$ Full embedded interrogative clause[7] (*which thing happened*)
f. **S**$_{\text{[-tense]}}$ Tenseless embedded clause (*something happen*)

For example, (8) gives the item instantiating the pair (*think, NP _ed that S*).[8]

(8) Someone thought that something happened.

[6] See Appendix A for an explicit mapping of the abstract frames in (5) to their corresponding instantiated frames.

[7] White et al. only use adjunct questions to avoid free relative readings. We instead opt for D-linked WH questions, though for the same reason.

[8] The *something* instantiation of NP is only used for NPs in object position of a simple transitive (*NP _ NP*) and second object position in a double object construction (*NP _ NP NP*).

From the resulting set of 30 verbs × 46 frames = 1,380 items, we constructed 23 lists of 60 items, constrained such that each verb occurred exactly twice in each list (always with a distinct frame) and each frame occurred between one and two times in each list.[9] Each item consists of a sentence paired with an ordinal acceptability judgment using a 7-point ordinal (or Likert) scale value. Subjects were presented with items in the browser, following instructions and other introductory material. Figure 4 gives the instructions, which were the same for the pilot and the full experiment, modulo the number of items mentioned.

## 3.2 Participants

We recruited 115 unique participants = 23 list × 5 participants per list through Amazon Mechanical Turk. All participants reported speaking American English as their native language.

## 3.3 Predictions

To measure interannotator agreement, White et al. compute the Spearman rank correlation between the responses for each pair of participants that did the same list and report a mean correlation of 0.64. This measure is not comparable to ours, however, since our items were specifically designed to block lexical information from being used in the acceptability judgments. This means that participants in principle have more potential interpretations to consider when making a judgment; and depending on the variability in acceptability of such interpretations, we expect no less (and possibly more) variability in responses to bleached items. Thus, we expect lower agreement. To assess how much lower we should expect, we attempt to factor out the variability across items that instantiate a particular verb-frame pair. To derive this more comparable measure, we simulate the amount of agreement we would expect, assuming they had used a method like ours.

First, we fit an ordinal (linked logit) mixed effects model to the ratings from White et al.'s data, with fixed effects for VERB, FRAME, and their interaction and random unconstrained cutpoints for each participant (see

[9] This distribution is necessary because there is no way to enforce that each verb occur an equal number of time and that each frame occur an equal number of times without having extremely small or extremely large lists. And because our aim is to match the frames used by White et al. as closely as possible, it would be problematic to manipulate the number of frames to make this constraint feasible.

Appendix B for details). We then use this model to simulate how each participant from their experiment would respond to each verb-frame pair in their experiment by (i) using the ordinal model to produce a predicted probability distribution over the ordinal scale ratings for each participant and item; (ii) sampling once from each of those distributions; and (iii) computing the Spearman correlation between responses given by all pairs of simulated participants. We repeat this simulation 999 times, computing the mean agreement each time.[10] This yields a mean correlation of 0.516 (95% CI: [0.511, 0.521]) across all simulations.

## 3.4 Results

We are concerned with two sorts of results in this validation: (i) interannotator agreement among participants in the validation compared to interannotator agreement in White et al.'s (2018a) data; and (ii) agreement in the aggregated ratings for verb-frame pairs.

To measure interannotator agreement, we compute the Spearman rank correlation between the responses for each pair of participants that did the same list. This yields a mean correlation of 0.528 (95% CI: [0.509, 0.545]). Thus, the level of interannotator agreement we observe is exactly what we expect given data collected under a more standard methodology.

Next, we turn to agreement between the average ratings for each verb-frame pair computed from each dataset. To compute these average ratings, we fit the ordinal mixed model described in the last subsection to each dataset separately and then compute the predicted real-valued acceptability for each verb-frame pair.[11]

Figure 2 plots the Spearman rank correlation between these normalized, real-valued verb-frame acceptability by frame (across verbs), and Figure 1 plots the same correlation by verb (across frames).[12] In both plots, the dashed line shows the mean interannotator agreement, and error bars show 95% confidence intervals.

[10] Throughout the remainder of the paper, all confidence intervals are computed using non-parametric bootstraps with 999 replicates.

[11] This procedure is analogous to $z$-scoring the ratings by participant and then computing the average $z$-scored rating for each item. As shown by White et al. (2018a), the method used here better models how participants actually make acceptability judgments. See Appendices B and C for further details on the conceptual and empirical relationship between $z$-scoring and the ordinal model-based normalization we use.

[12] An anonymous reviewer asks why we use Spearman correlation instead of a procedure wherein we bin the ordinal acceptability responses into *unacceptable* and *acceptable* bins and then use a measure like Cohen's $\kappa$. This latter procedure is undesirable for two reasons. First, it unnecessarily increases the researcher degrees-of-freedom by allowing the researcher to choose how to do the binning, which in turn allows the researcher to optimize

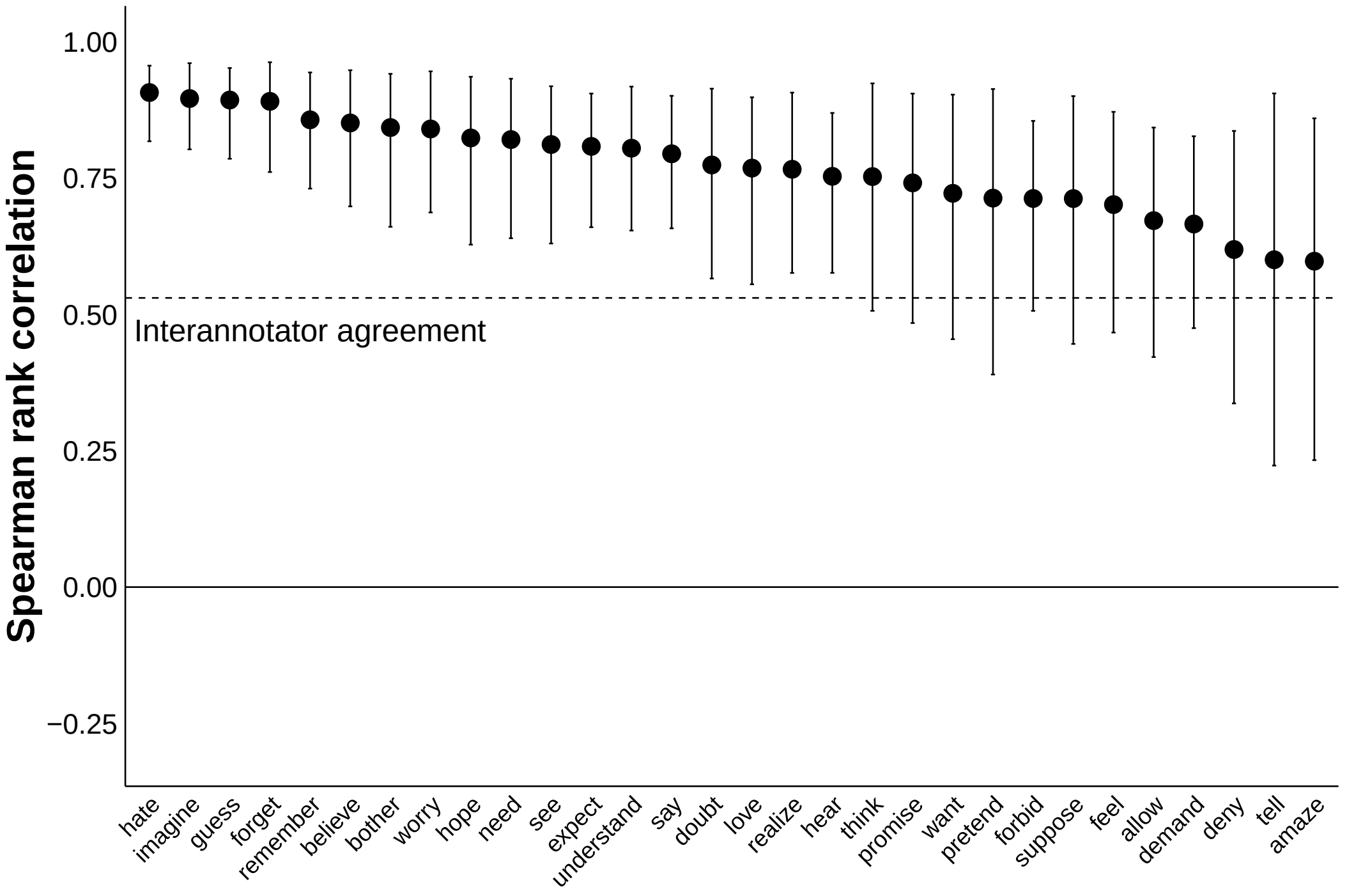

**Figure 1:** Correlation by verb between mean normalized verb-frame acceptability in White et al.'s (2018a) data and our replication.

In Figure 1, we see that all verbs show correlations above the mean interannotator agreement. This suggests that as a measure of verbs' syntactic distributions, our data are encoding essentially the same distributional information that White et al.'s data are, and there are no substantial differences tied directly to the two verbs.

In Figure 2, we see that most frames show average correlations close to or above the mean interannotator agreement, and we also take these cases to involve no substantial differences between the two experiments tied to those frames. There are five frames that do not show a correlation that is significantly different from zero: *NP _ NP*, *NP _ NP NP*, *It _ NP that S*, *It _ NP* the binning method to make the agreement look as good as possible. Spearman correlation does not have this problem, since it does not requiring a binning step. Second, such a binning procedure throws away important information about the relative acceptability of different items, whereas Spearman correlation allows us to incorporate this information holistically into our measure.

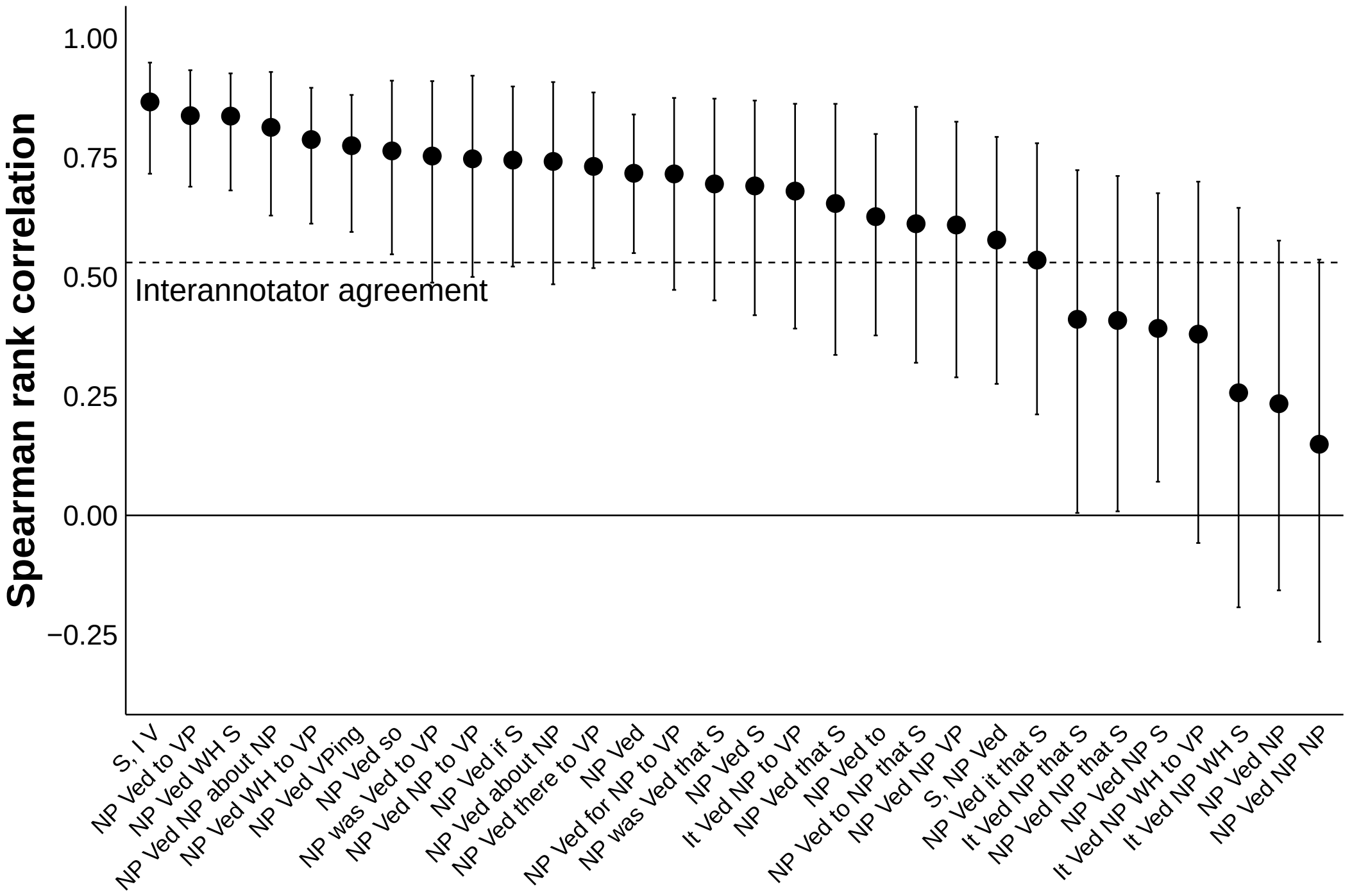

**Figure 2:** Correlation by frame between mean normalized verb-frame acceptability in White et al.'s (2018a) data and replication.

*WH S*, and *It _ NP WH to VP*. We therefore discuss potential explanations for the reasons why these frames differ between the two experiments.

For *NP _ NP* and *NP _ NP NP*, it seems likely that the disagreement arises because White et al.'s instantiations of those frames only ever include object NPs that denote concrete inanimate entities that cannot be straightforwardly associated with propositional content—e.g. cups, tables, bottles. In contrast, the inanimate indefinites we use could denote either contentful inanimates or contentless ones. This is likely the cause of higher acceptabilities observed for predicates like *believe* and *tell*. Compare (9a) and (9b) with (10a) and (10b).

(9) a. #I believed the table.
b. #I told her the table.[13]

(10) a. Someone believed something.
b. Someone told someone something.

[13] Note that these ditransitive frames are licit with verbs such as *allow*, *deny*, and *forbid*.

Since we aim to factor out effects due to lexical items, this is a point in favor of our method.

For *It _ NP that S*, *It _ NP WH S*, and *It _ NP WH to VP*, we suspect that the low agreement stems from inherent variability in the judgments for items with expletive subjects. One reason this may arise—pointed out by White et al.—is that *it* can be read referentially in these frames, and thus participants' judgments might vary widely depending on their interpretation of *it*. This predicts that expletive subject frames should show lower agreement on average, which appears to be the case.

Regardless of its source, the existence of this low agreement for expletive subject frames suggests that we should be wary of including such frames in a large-scale experiment like the one we report on in this paper. Nonetheless, we would still like to capture information about whether a verb allows expletive subjects. We discuss our approach to this below.

## 3.5 Discussion

The results reported above suggest that the bleaching method is a promising way to avoid item effects in acceptability judgment tasks for selectional patterns. Since it involves an extremely simple generation strategy, it therefore is also a promising way of scaling standard acceptability judgment tasks to entire subregions of the lexicon. But even with bleaching, one must thread the needle between useful referential ambiguities—such as the one between entity and propositional reference introduced by the use of *something*—and plausibly syntactic ambiguities that introduce variability into the judgments—such as the one between referential and expletive *it* pointed out above.

We address the particular issue of expletive *it* in our lexicon-scale annotation by using an alternative set of structures for capturing the acceptability of predicates that occur with expletive *it*: at least those predicates that take experiencer objects. Our approach, which we describe in the next section, is to use passivized version of the expletive object frames; compare (11a) and (11b).

(11) a. It amazed someone that something happened.
b. Someone was amazed that something happened.

This approach introduces some amount of ambiguity—we don't know whether a verb that is acceptabile in contexts like (11b) takes contentful or expletive

subjects—but this ambiguity is resolvable by looking at the acceptability of that verb in contexts such as (12).

(12) ???Someone amazed someone that something happened.

That is, a verb is licensed in an expletive subject frame if it is licensed in a passive transitive frame, and not licensed in a non-passive ditransitive frame. In our large-scale experiment, we present an expanded set of frames using this manipulation.

# 4 The MegaAcceptability Dataset

The main goal of collecting our large-scale acceptability judgment dataset is to obtain a single normalized acceptability score for every clause-embedding verb in the English lexicon, along with an estimate of the variability in judgments for that item. We discuss our data collection method and how we derive these estimates here. In Section 5, we describe experiments that attempt to predict these normalized acceptabilities from frequency data.

## 4.1 *Materials and data collection method*

To scale up the materials, we selected a set of frames, a set of verbs, and automated a method of constructing bleached sentences for every member of the cartesian product of the two.[14]

For verb selection, we attempted to exhaustively select every verb in English which could take a clause of some kind. First, we took the union of several lists of clause-embedding verbs collected in previous work (Hacquard & Wellwood 2012; Anand & Hacquard 2013; 2014; Rawlins 2013; White et al. 2014) as a 'seed' set. Helpfully, a range of existing lists were already aggregated in Rawlins (2013) and constituted the bulk of the seed. This gave us about 500 verbs.

We then searched in VerbNet (Kipper-Schuler 2005)—a database that is, in large part, directly derived from the verb classes in Levin 1993—to find all verbs in all VerbNet classes that any of the seed verbs were present in, with a hand-filtering pass to remove obvious errors—e.g. cooking verbs.

To pick the set of frames, we first collected a set of five basic syntactic features that are believed to be relevant to selectional patterns, and selected either all or the most frequent values for these features. In this case, we did not aim for full exhaustivity, but rather to get as big a sample as possible within constraints imposed by the already large experiment.

[14] These materials were first described in White & Rawlins 2016. See Appendix A for an explicit mapping of the abstract frames to their corresponding instantiated frames in Figure 3.

For example, for prepositional phrases, we consider only the prepositions *to* and *about*, though many other prepositional markers, such as *of* and *from*, may be relevant to the ultimate question of how to represent selectional patterns. For embedded constituent interrogatives, we chose to use an embedded D-linked WH-phrase (*which thing*) in order to maximize acceptability. To these features, we added passivization—in order to handle expletive subjects as described above—and two more idiosyncratic frame manipulations: declarative slifting (Ross 1973), and the proform *so* (Ross 1972; Hankamer & Sag 1976).

(13) a. **Complementizer**: *∅, that, for, whether, which thing*
b. **Embedded tense**: *past, future, infinitival, present particple, bare*
c. **Matrix object count**: 0, 1, 2
(i) **so-frame manipulation**: add *so* to 0-object frames
d. **Matrix PP**: *∅, to, about*
e. **Embedded subject?**: true, false
f. **Passivized verb?**: true, false
g. **Slifting manipulation**: 'Something happened, I ___'

The frame instantiations are shown in Figure 3. As can be seen in this figure, the bleaching manipulation we applied is identical to that applied in the validation experiment reported in Section 3.

Each item consisted of a sentence constructed using the bleaching method from a verb and a frame as described above, paired with an ordinal acceptability judgment using a 1-7 point ordinal (Likert) scale. From the base set of verbs and frames described so far, we constructed 1,000 lists of 50 items each. For this experiment (in contrast to the pilot experiment reported above), each frame and each verb appear at most once in each list. Each list was presented to the subject in a browser via Amazon Mechanical Turk as a single page, with subsequent items reached by scrolling. A sample view of the first four items of a list are provided in 5.

Each list was provided to participants as a Human Intelligence Task (HIT) in Mechanical Turk. Participants were presented with the instructions and training items depicted in Figure 4, followed by several demographics questions (including native language), IRB information, and finally the items from their list. Each item involved rating the acceptability of one of the constructed sentences used a 1-7 point ordinal (Likert) scale. In or-der to submit each HIT, participants needed to check a box indicating their consent to participate in the study, as part of the IRB information.

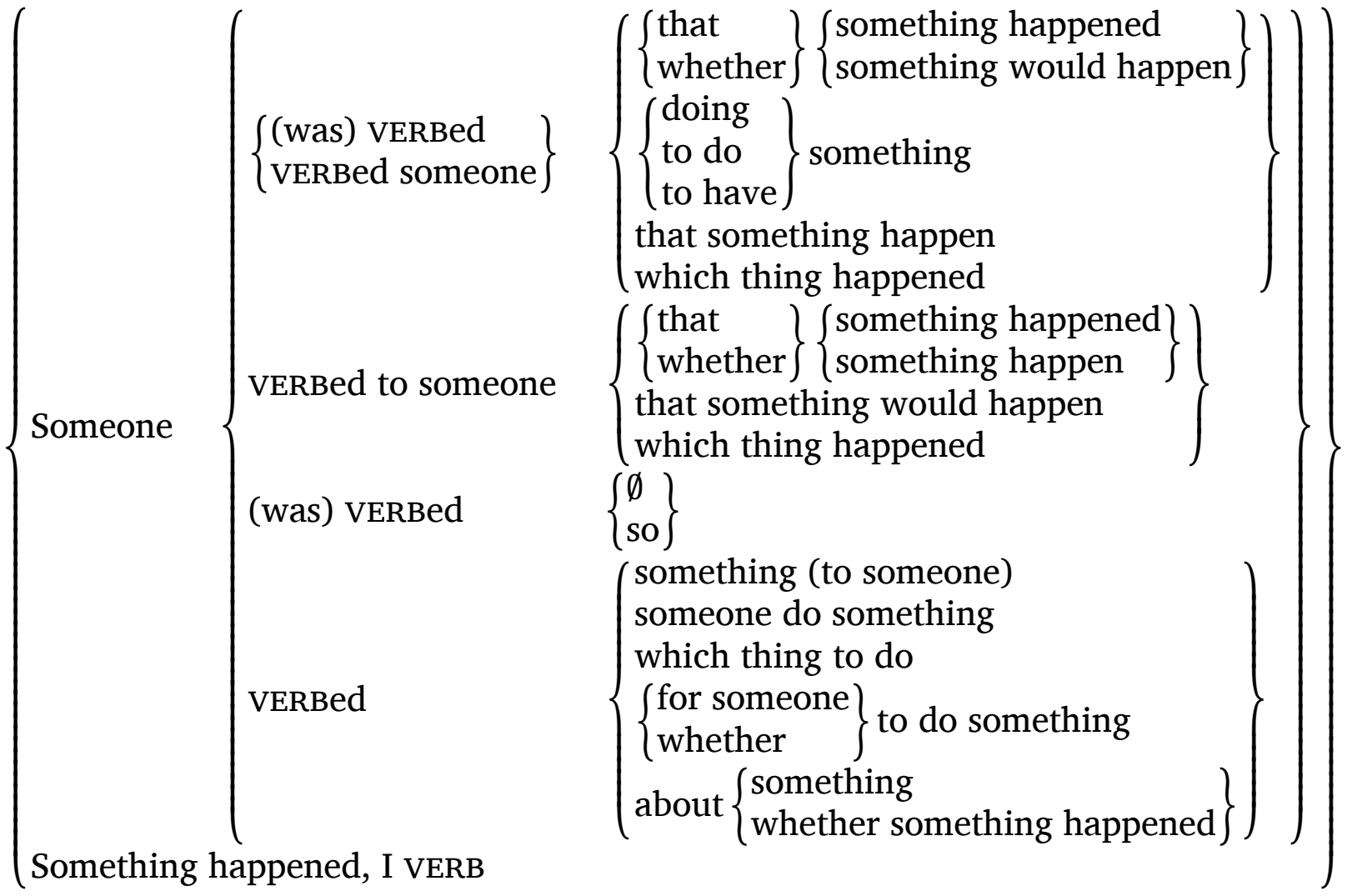

**Figure 3:** All frame instantiations in the MegaAcceptability dataset (White & Rawlins 2016, Figure 4).

## 4.2 *Participants*

727 unique participants were recruited through Amazon Mechanical Turk to rate sentences in the 1,000 lists of 50. Participants were allowed to respond to as many unique lists as they liked.[15] No one participant was allowed to rate the same list twice, and each list was rated by five unique participants,

[15] In allowing participants to respond to multiple lists, we were attempting to balance three pressures. First, ideally, we would ask participants to respond to as many sentences as possible because an increase in the numbers of responses from any particular participant allows us to better normalize that participant's ratings relative to other participants (via the prior distribution on random effects in the mixed effects model-based normalizer). Second, working against this first concern, we attempted to have as many distinct participants as possible to enable better estimation of common patterns in participants' response behavior—also, helping us to better normalize judgments, especially for participants who gave fewer responses (again, via the prior distribution on random effects). And third, we did not know whether or not it would be feasible to recruit 5,000 distinct participants, since our experiment is substantially larger than others that do not allow repeat participation. Further, most crowd-sourcing annotation tasks of similar size to ours—e.g. those found in the NLP literature (see Callison-Burch 2019: and references therein)—allow repeat partic-

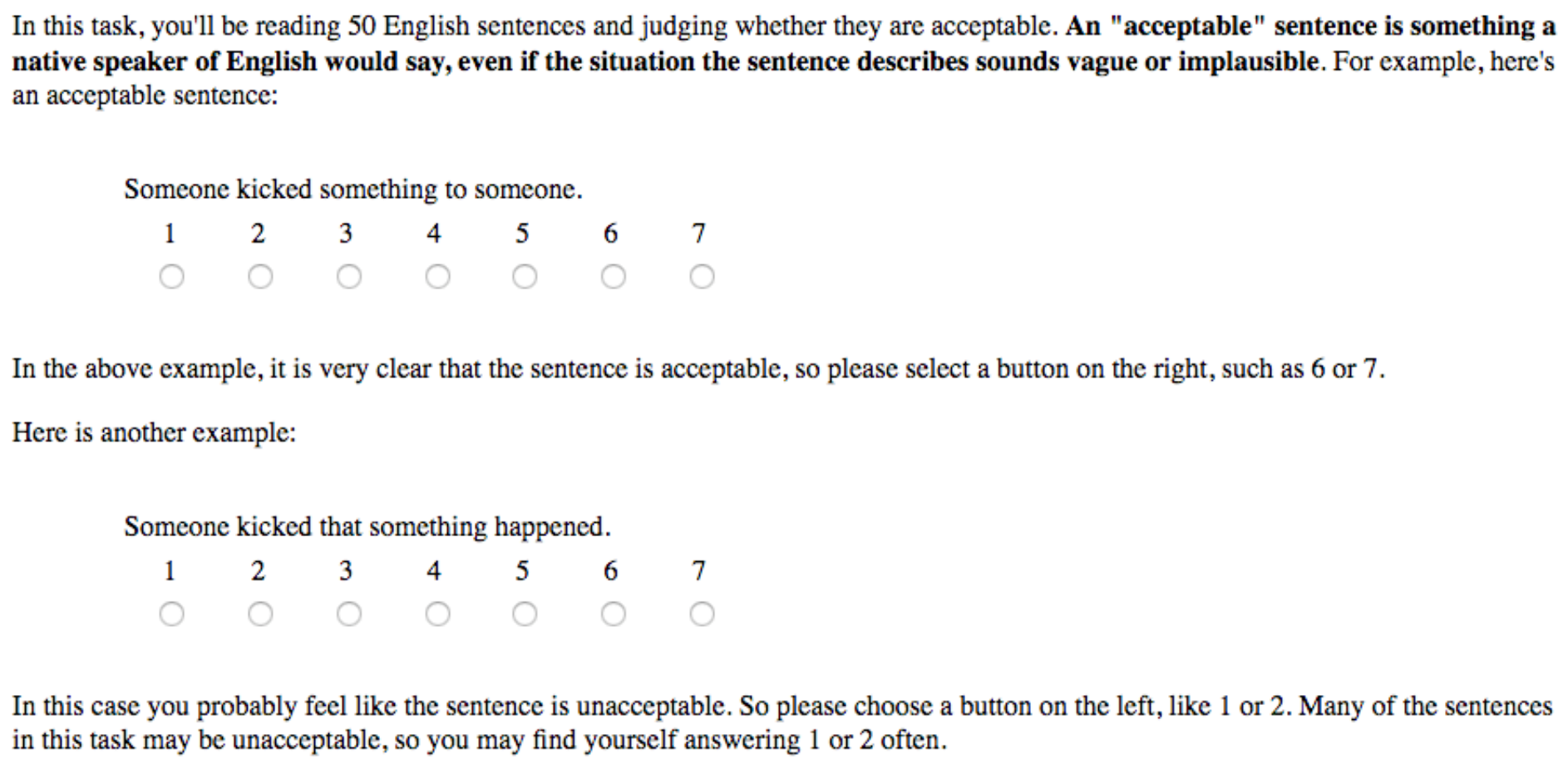
In this task, you'll be reading 50 English sentences and judging whether they are acceptable. **An "acceptable" sentence is something a native speaker of English would say, even if the situation the sentence describes sounds vague or implausible**. For example, here's an acceptable sentence:

Someone kicked something to someone.

1 2 3 4 5 6 7

In the above example, it is very clear that the sentence is acceptable, so please select a button on the right, such as 6 or 7.

Here is another example:

Someone kicked that something happened.

1 2 3 4 5 6 7

In this case you probably feel like the sentence is unacceptable. So please choose a button on the left, like 1 or 2. Many of the sentences in this task may be unacceptable, so you may find yourself answering 1 or 2 often.

**Figure 4:** Instructions and example items for each list.

leading to five unique ratings per item. Each participant responded to a median of four lists (mean: 6.9, min: 1, max: 56).

Four participants reported being native speakers of a language other than English. These participants' responses were removed from the dataset prior to analysis, for a loss of 600 responses total (∼0.2% of the data). None of these participants rated the same list.

## 4.3 *Response Normalization*

We use a slightly modified form of the ordinal model-based normalization procedure described in Section 3 to produce two pieces of information associated with each verb-frame pair: a real-valued acceptability value (more positive is more acceptable) and the mean (log-)likelihood associated with all acceptability judgments for a particular item.[16] The second score can be viewed as a measure of variability in the judgments: the lower this likelihood score is, the higher the variability in ordinal responses to a particular verb-frame.

As an example of what these two measures look like and their relationship to the original ratings, Figure 6 plots the mean ratings for the *NP ___ed* ipation. Thus, it was not clear whether 5,000 unique annotators would actually complete our task, which is much more involved than most large-scale crowd-sourcing tasks wherein the task takes on the order of seconds. Further, as laid out in the text, we have taken pains to correct for any potential annotation biases arising from allowing repeat participation.

[16] A full specification of this procedure, including a comparison to alternative methods for aggregating participants' responses to particular sentences, can be found in Appendix C.

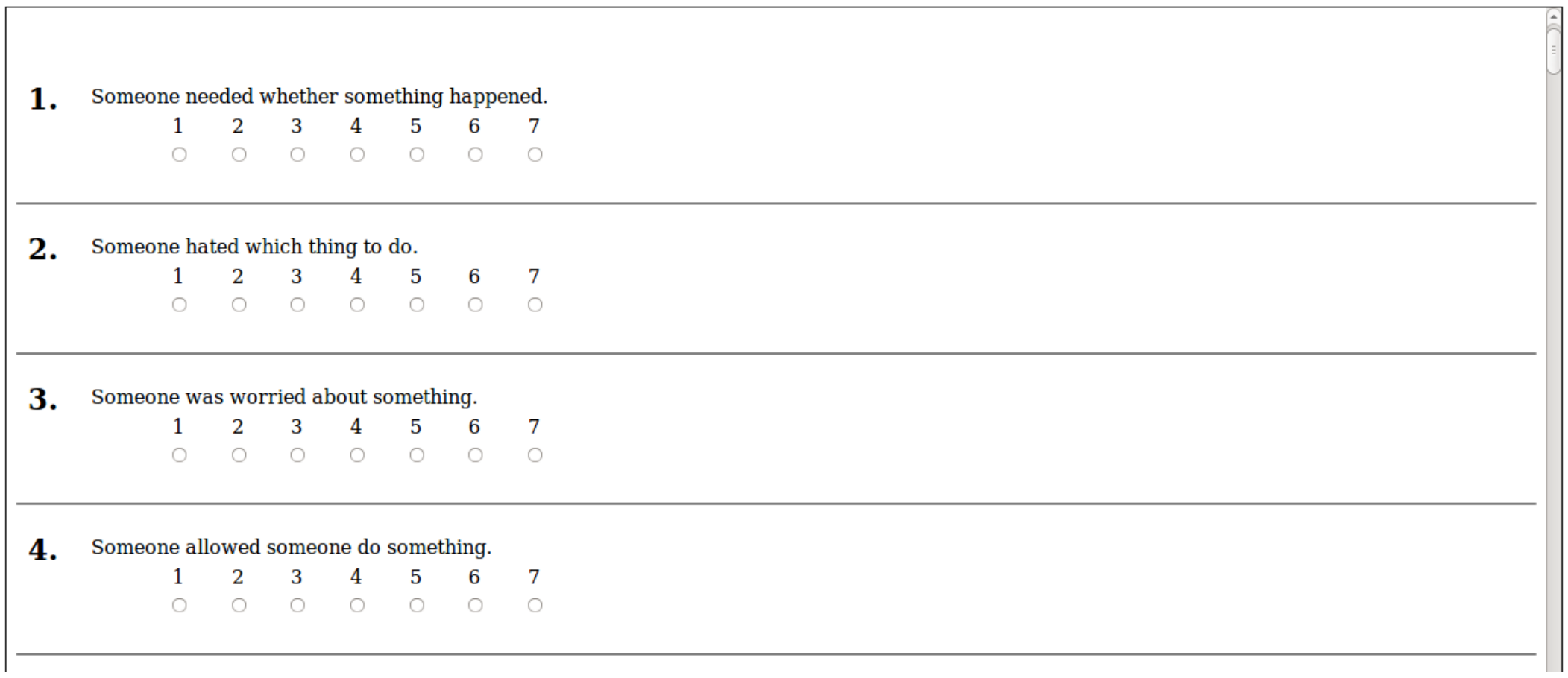

**Figure 5:** Four items from a sample list.

*that S* and *NP ___ed NP that S* frames (treating the ordinal ratings as though they were interval data), and Figure 7 plots the normalized acceptability scores for those same frames, where more to the right (top) means higher normalized acceptability and more to the left (bottom) means lower acceptability.[17] Each point is a verb and only a subset of points are labeled. In Figure 7 smaller labels and grayer points correspond to higher mean variability—i.e. lower likelihood score.

As one would expect, verbs like *think*, *assume*, *discover*, and *notice* are very acceptable in the *NP ___ed that S* frame but quite bad in the *NP ___ed NP that S* frame, and there is little variability in these ratings. In contrast, verbs like *tell*, *remind*, and *notify* are very good in the *NP ___ed NP that S* frame but middling in the *NP ___ed that S* frame, with more variability in their ratings. This variability is due in particular to the judgments for the *NP ___ed that S* frame, suggesting that some participants are okay with dropping the object while others are not. This contrasts with a verb like *persuade*, for which participants are more unified in their dislike of object drop.

## 4.4 *Reliability*

In Section 3, we compared the normalized acceptability obtained from White et al.'s (2018a) experiment and our replication that used the bleaching method. We cannot derive similar agreement estimates here because we only have a single dataset and because the lists were built in such a way that participants only saw one verb and one frame; it is therefore not possible to assess the correlations within a particular frame.

[17] The axes in Figure 7 are not labeled because these normalized scores do not have inherent meaning beyond measuring the relative acceptability of a verb in a frame.

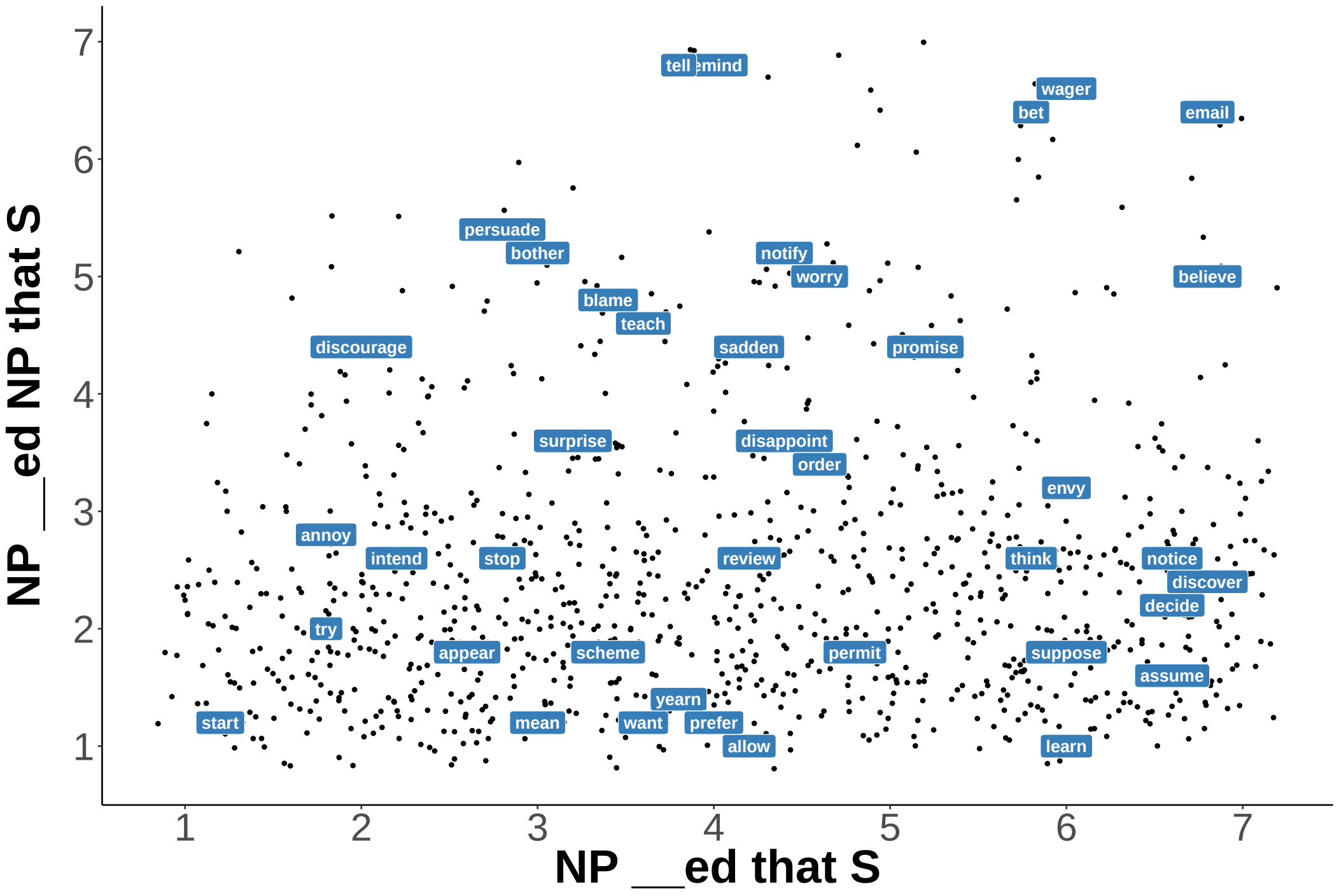

**Figure 6:** Mean of raw ratings for two frames. Each point is a verb (jittered to mitigate overplotting) and only a subset of points are labeled.

However, because we do have estimates of the variability in judgments for each verb-frame pair, we can get a sense for how much agreement there is within judgments for a particular frame by taking the mean of the above-defined variability scores for each frame, across verbs. Remember that these variability scores are just mean likelihood values and that higher likelihood values correspond to lower variability (see Appendix B). Figure 8 plots these means in terms of probabilities. A value of 0.14 ($\frac{1}{7}$) is the lowest possible probability—roughly corresponding to each participant giving equally spaced values along the ordinal scale. We see that nearly all frames fall within a narrow band between 0.3 and 0.5, suggesting that no frame shows particularly high disagreement—in contrast to what we saw in Section 3 for the expletive subject and NP direct object frames. Further, our replacements for the expletive subject frames—the passivized frames—do not show systematically lower variability, suggesting that our approach was successful.

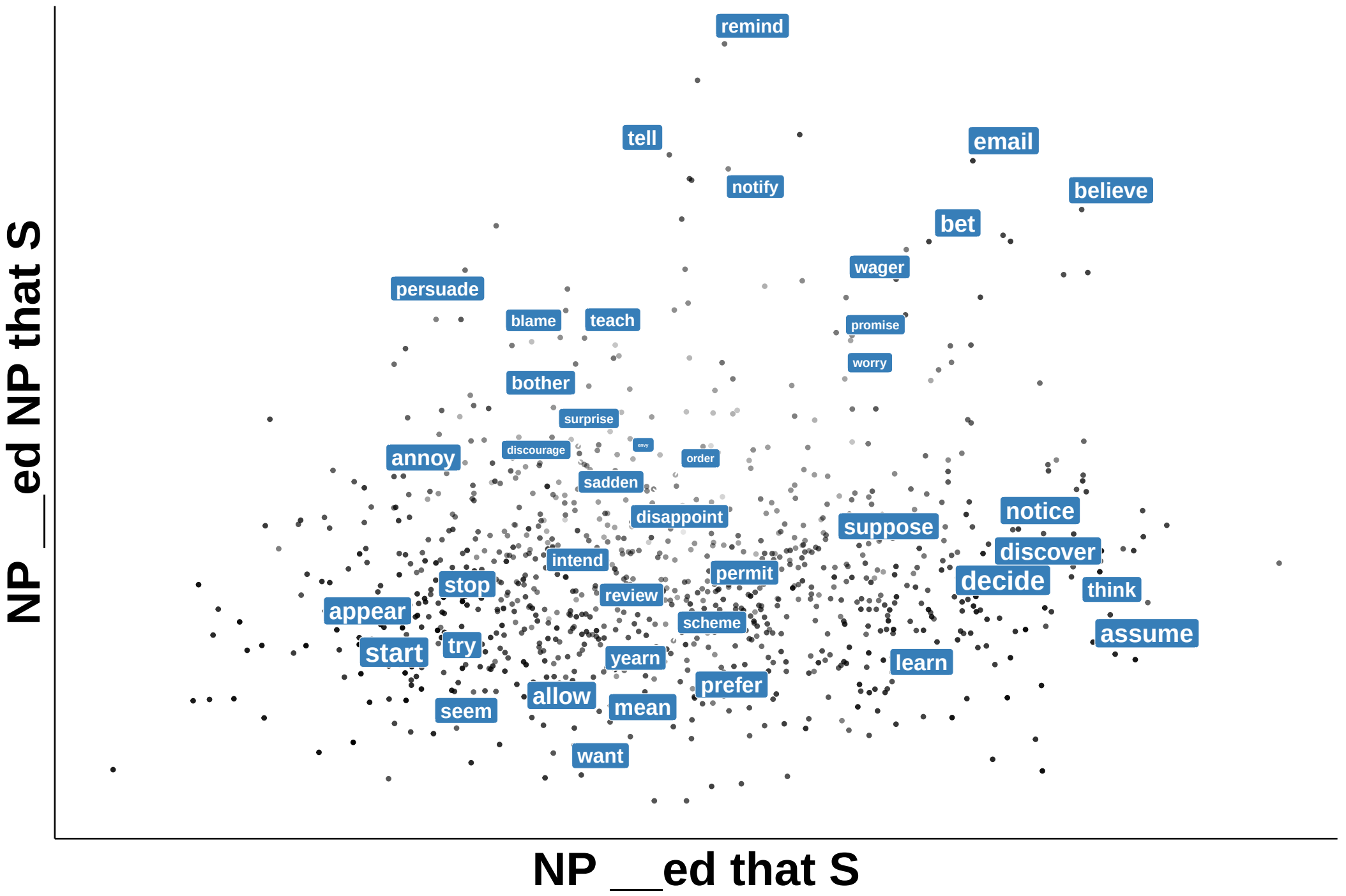

**Figure 7:** Normalized judgments for two frames. Each point is a verb and only a subset of points are labeled. Smaller labels and grayer points correspond to higher mean variability.

(See White & Rawlins (2016) for further discussion of how this data might be used or validated in a formal semantics context.)

## 5 Relating Frequency and Acceptability

We now turn to the main question of this paper: to what extent can a verb's subcategorization behavior be predicted directly from the frequency with which it occurs in different syntactic structures in text? To obtain our measure of frequency, we use the VALEX dataset, which is the largest publicly available dataset of subcategorization frame frequencies (Korhonen et al. 2006).[18] VALEX is built from over 900 million words of text and contains 163 subcategorization frame types, described in Briscoe & Carroll 1997, and over 6,000 verbs, 958 of which are shared with the MegaAcceptability dataset. (The verbs that are missing tend to be particle verbs, as these are not treated as separate verbs by VALEX.)

[18] VALEX is available at https://ilexir.co.uk/valex/. We use the raw counts provided with the data, since all other counts involve some amount of smoothing and/or filtering.

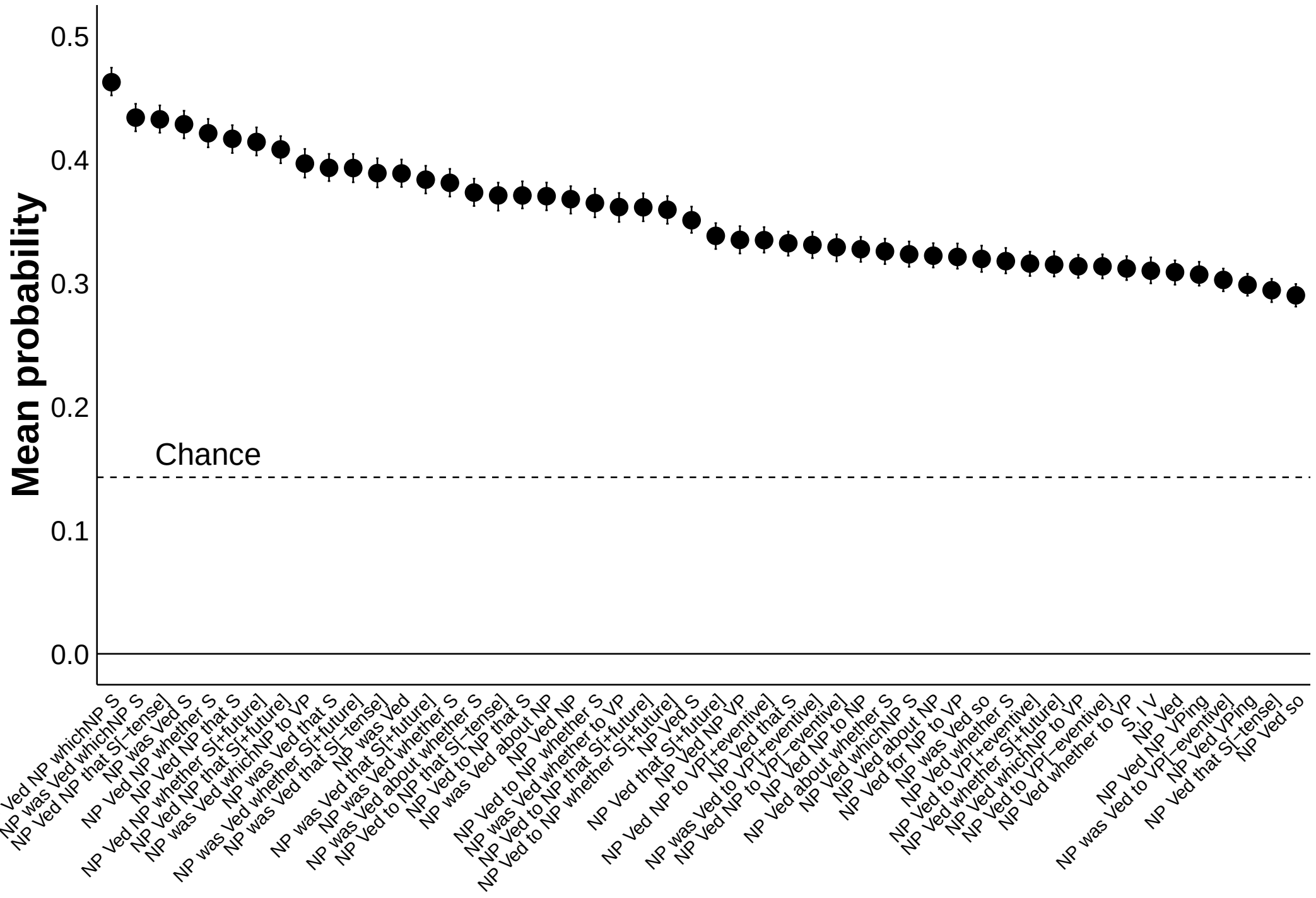

**Figure 8:** Means of verb-frame variability judgments, for each frame. Higher probability means lower variability.

One obstacle to using any frequency dataset for predicting acceptability is that we must determine the importance of any particular verb-frame co-occurrence in the context of its entire distribution. For example, observing a high-frequency verb like *think* once or twice in a ditransitive frame should make us less certain that *think* is highly acceptable in that frame than observing a lower frequency verb like *begrudge*. This situation is delicate, though, since it requires the specification of some *frequency normalization model* for processing the raw frequency data and we do not want to introduce too much inductive bias at this stage to ensure that we are in fact testing the predictability of acceptability directly from frequency.

To thread this needle, we consider only normalization models that represent verbs' distributions directly in terms of the original subcategorization frames—as opposed to some set of latent syntactic or semantic factors—while accounting for the importance of a particular observation in the context of those distributions. We then use each of these representations as predictors in a linear model of the normalized acceptability judgments described in the last section.

The use of a linear model for this purpose is important so as not to introduce any further inductive bias—as, e.g., the use of a kernelized support vector machine or multi-layer perceptron might. Further, since linear models learn linear functions from one representation to another and since linear functions are all and only the homomorphisms (structure-preserving mappings) between those representations, this setup allows us to make stronger conclusion about the character of the relationship—specifically, whether or not the (normalized) frequency distributions and acceptability are homomorphic (structurally similar).

## 5.1 Normalization Models

We consider two probabilistic models and two information theoretic models of subcategorization frame frequency distributions. Our aim in looking at multiple models is less to compare their relative performance, and more to give the frequency information the best chance at explaining acceptability.

### 5.1.1 Probabilistic Models

The first probabilistic model we consider models the conditional probability $\mathbb{P}(f \mid v)$ of seeing a particular frame $f$ given a particular verb $v$ as a Categorical distribution with parameters (probabilities) $\boldsymbol{\theta}_v$.

$$\mathbb{P}(f \mid v) = \mathbb{P}(f \mid \boldsymbol{\theta}_v) = \text{Categorical}(f; \boldsymbol{\theta}_v) = \theta_{vf}$$

We use the frequencies $c_{vf}$ for each verb $v$ and frame $f$ to compute the posterior probability $\mathbb{P}(\boldsymbol{\theta}_v \mid \boldsymbol{c}_v)$ of $\boldsymbol{\theta}_v$ under the assumptions (i) that the verb-frame pairs are sampled independently; and (ii) that the prior probability $\mathbb{P}(\boldsymbol{\theta}_v; \boldsymbol{\alpha})$ is given by a Dirichlet distribution with parameter $\boldsymbol{\alpha}$.

$$\mathbb{P}(\boldsymbol{\theta}_v \mid \boldsymbol{c}_v; \boldsymbol{\alpha}) \propto \mathbb{P}(\boldsymbol{\theta}_v; \boldsymbol{\alpha}) \prod_f \mathbb{P}(f \mid \boldsymbol{\theta}_v)^{c_{vf}}$$

We then use the most likely probabilities $\hat{\boldsymbol{\theta}}_v$—i.e. the *Maximum A Posteriori* (MAP) estimate—for each verb as our representation of a verb's distribution—i.e. $\hat{\boldsymbol{\theta}}_v$ is what we use to predict a verb's acceptability. When $\boldsymbol{\alpha}$ is a constant positive vector $(\lambda + 1)\mathbf{1}_{N_F}$ (where $N_F$ is the number of frames), this turns out to be equivalent to standard add-$\lambda$ smoothing.

$$\hat{\boldsymbol{\theta}}_v = \arg_{\boldsymbol{\theta}_v} \max \mathbb{P}(\boldsymbol{\theta}_v \mid \boldsymbol{c}_v; \boldsymbol{\alpha} = (\lambda + 1)\mathbf{1}_{N_F}) = \left[ \frac{c_{v1} + \lambda}{\sum_i c_{vi} + \lambda}, \frac{c_{v2} + \lambda}{\sum_i c_{vi} + \lambda}, \dots \right]$$

Thus, a special case of this model ($\lambda = 0$) just involves dividing a verb's frequency in a frame by the verb's frequency across all frames. Regardless of the setting of $\lambda$, the verb's frequency representation always sums to 1 in this model, and $\lambda > 0$ enforces that, even if a verb-frame pair hasn't been seen, there is still some probability that it might be seen in the future, with the amount of probability assigned to those unseen verb-frame pairs dependent on the size of $\lambda$ (see Jurafsky & Martin 2009: Ch. 4).

The second probabilistic model attempts to directly extract acceptability from the frequency data by finding, for each verb-frame pair, a probability that that verb is acceptable in that frame (White 2015; White et al. 2017a). In this model, the conditional probability of seeing a particular frame $f$ with some frequency $c_{vf}$ given a verb $v$ is assumed to have a negative binomial distribution with probability $\pi_{vf}$ (the probability that $v$ is acceptable in $f$) and rate $r_v$ (roughly, corresponding to the overall frequency of the verb $v$).

$$\mathbb{P}(c_{vf} \mid v) = \text{NegBin}(c_{vf}; \pi_{vf}, r_v) = \binom{c_{vf} + r_v - 1}{c_{vf}} (1 - \pi_{vf})^{r_v} \pi_{vf}^{c_{vf}}$$

This distribution is a natural choice both (i) because it is known to be a good model of similar kinds of count data (Church & Gale 1995); and (ii) because the parameters themselves have natural interpretations: (a) the parameter $\pi_{vf}$ can be viewed as a probability of acceptability: when it is close to one, we expect to see more instances of a frame with a verb (though there is a non-zero probability of seeing it rarely); when it is close to zero, we expect to see fewer; and (b) the verb's rate parameter $r_v$ roughly controls its overall frequency. Thus, unlike for the Dirichlet-Categorical model, we straightforwardly separate our knowledge of competence ($\pi_{vf}$) from our knowledge of frequency ($r_v$; for further discussion, see White 2015). This is particularly evident from the fact that, the Dirichlet-Categorical model's representation must sum to 1—thus, telling us the probability of *seeing* a particular verb-frame pair—while the Beta-Binomial model's representation does not have such a requirement: a verb $v$ can be acceptable in more than one frame $f$, represented by $\pi_{vf}$ being near 1 for each of those frames; information about the probability of actually *seeing* the verb-frame pair is largely factored into $r_v$.

As with the $\boldsymbol{\theta}_v$ parameter of the Categorical-Dirichlet model, we aim to find the most likely pairing $\hat{\boldsymbol{\pi}}_v, \hat{r}_v$ for each verb $v$, given the counts $\mathbf{c}_v$. We assume that the prior probability $\mathbb{P}(\pi_{vf}; \beta_1, \beta_2)$ in this case is Beta distributed with parameters $\beta_1 = \beta_2$—henceforth, referred to via $\gamma = \beta_1 - 1 = \beta_2 - 1$.[19] We assume an improper (uniform) prior for $r_v$.

$$(\hat{\pi}_{vf}, \hat{r}_v) = \arg_{\pi_{vf}, r_v} \max \mathbb{P}(\pi_{vf}, r_v \mid \boldsymbol{c}_v; \beta_1, \beta_2)$$

Unlike for $\hat{\boldsymbol{\theta}}_v$, this MAP estimate for $\hat{\boldsymbol{\pi}}_v, \hat{r}_v$ cannot be computed using a closed form and so we use gradient descent to obtain it.

For both the Dirichlet-Categorical model and the Beta-Negative Binomial model, we refer to the hyperparameters $\lambda$, in $\boldsymbol{\alpha} = (\lambda + 1)\mathbf{1}_{N_F}$, and $\gamma$ as *smoothing parameters*. We consider multiple different settings of smoothing parameters in our experiments, described below.

### 5.1.2 Information Theoretic Models

The first information theoretic model we consider uses the *pointwise-mutual information* (PMI; Church & Hanks 1990) between a verb and a frame.

$$\text{PMI}(v, f) = \log \frac{\mathbb{P}(v, f)}{\mathbb{P}(v) \cdot \mathbb{P}(f)}$$

This quantity is commonly used to find *collocations*—i.e. common pairings of words or phrases (see Manning & Schütze 1999: Ch. 5). To compute this quantity, we assume the Dirichlet-Categorical model described above and obtain MAP estimates for the parameters of the joint distribution $\mathbb{P}(v, f)$. The parameters of the marginal distributions $\mathbb{P}(v)$ and $\mathbb{P}(f)$ can then be obtained from the joint. We estimate these distributions using the same definition of the smoothing parameter $\lambda$.

The second information theoretic model we consider uses the terms of the $G$ statistic, as described by Dunning (1993). This amounts to scaling the PMI by the relevant conditional probability—in our case, the probability of the frame given the verb—better controlling for relatively poorer probability estimates for low frequency verbs.

$$G(v, f) = \mathbb{P}(f \mid v) \cdot \text{PMI}(v, f)$$

As for PMI, we estimate these distributions using the same definition of the smoothing parameter $\lambda$.

[19] This is analogous to the Categorical-Dirichlet model: larger values of $\lambda$ encourage estimates of $\hat{\theta}_{vf}$ nearer $\frac{1}{N_F}$, and larger values of $\gamma$ encourage values of $\pi_{vf}$ nearer $\frac{1}{2}$.

## 5.2 Experiments

We compute MAP estimates for the parameters of the Dirichlet-Categorical model $\boldsymbol{\theta}_v$ and the Beta-Negative Binomial model $\pi_v, r_v$ with smoothing parameters $\lambda, \gamma \in \{0, 0.1, 0.2, 0.5, 1, 2, 5, 10, 20, 50\}$. We compute PMI and $G$ using MAP estimates based on the same settings of $\lambda$.

We regress the normalized acceptability judgments for each verb in each frame on each of these representations in a multivariate ridge regression—i.e. a linear regression with L2 regularization. To set the regularization parameter $\alpha \in \{0.01,\ 0.1,\ 0.2,\ 0.5,\ 1,\ 2,\ 5,\ 10\}$, we use a 10-fold cross-validation. To compute the generalizability of this model, we nest this cross-validation within another 10-fold cross-validation and compute the mean $R^2$ (variance explained) on the held out datasets in this outer cross-validation.

Similar to our reasoning for using multiple kinds of normalization models, our aim in using ridge regression with cross-validation as opposed to an unregularized regression fit to the entire dataset is to give each model the best chance at explaining acceptability for verb-frame pairs it has not seen. If we simply fit a linear regression for the whole dataset and then reported measures of fit on the same data, we could substantially overestimate the model's performance (see standard machine learning texts, such as Bishop 2006: Ch. 1-3). Importantly, though, the result of ridge regression just is a linear model, thus satisfying our goal of finding a homomorphic (structure-preserving) mapping from the frequency representation to the acceptability representation.

## 5.3 Results

Figure 9 shows the mean $R^2$ across the 10 cross-validation folds for each model and smoothing parameter. We see that the Beta-Negative Binomial model ($\gamma = 0.1$) is the best-performing model, with the PMI model ($\lambda = 5$) a close second. The Dirichlet-Categorical model ($\lambda = 0$) is a close third, performing only slightly (but reliably) worse. The $G$ model consistently does more poorly than the other models—possibly because it too strongly downweights the scores for low frequency verbs (unlike the other measures, which attempt to filter out frequency to some extent). But even for the best-performing models, the scores are quite low. This suggests that, while the joint frequency of a verb and a frame carries some information about the acceptability of that verb in that frame, it is far from enough to determine that acceptability.

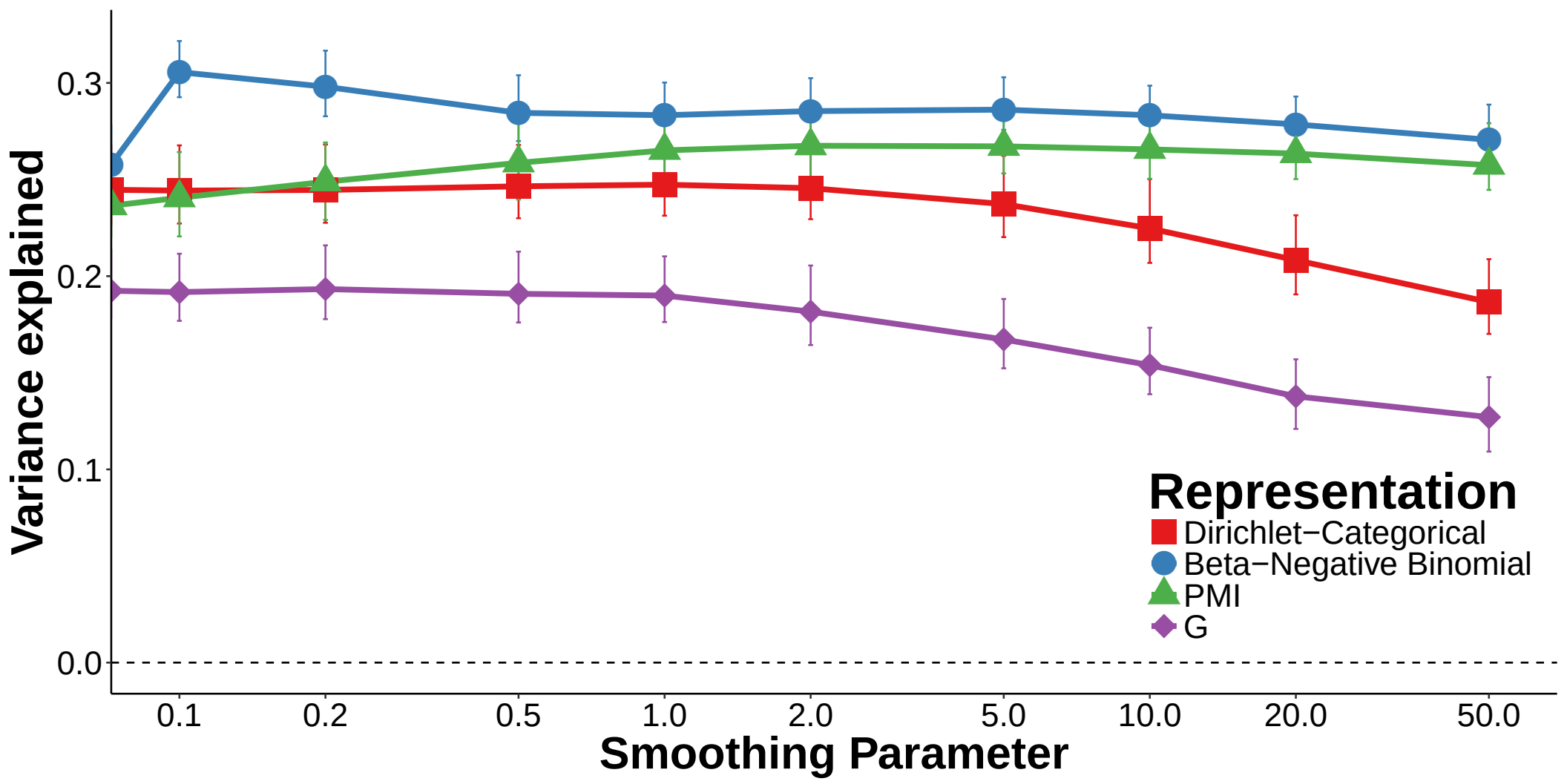

**Figure 9:** Mean variance explained in normalized acceptability judgments in 10-fold/10-fold nested cross-validation for each model and smoothing parameter.

One question that arises here is whether this poor performance is due to verb-frame pairs that received highly variable judgments. In this case, we should expect a positive correlation between judgment variability (as defined in Section 4) and models' absolute error. We test this hypothesis using our best-performing model's absolute error on the held-out data for each fold of the cross-validation. Rather, than finding a positive correlation, we find a weak (but reliable) negative Spearman rank correlation of -0.192 (95% CI = [-0.200, -0.184]).

This suggests that highly variable judgments are actually slightly easier to predict than less variable judgments. One reason this may come about is that more variable judgments tend to have normalized acceptabilities near the center of the acceptability scale, which arises from the fact that high variability is a consequence of extreme responses from participants that average out to the middle of the scale, as can be seen in Figure 10. This means that if a model incorrectly predicts a very high or very low acceptability score it will tend to be less wrong for the highly variable predicates in the middle of the scale than for predicates that participants were more certain about.

A similar question arises with respect to frequency: the poor performance we observe could be due to poor estimates for the distributions of low frequency verbs. In this case, we should expect a negative correla-tion between verb frequency and models' absolute error. We again test this hypothesis using our best-performing model's absolute error on the held-out data for each fold of the cross-validation. We instead find a reliably positive Spearman rank correlation here, though it is extremely weak 0.021 (95% CI = [0.011, 0.029]). This suggests that poor estimates of verbs' distributions—at least their frequency distributions—is not the cause of our models' poor performance.

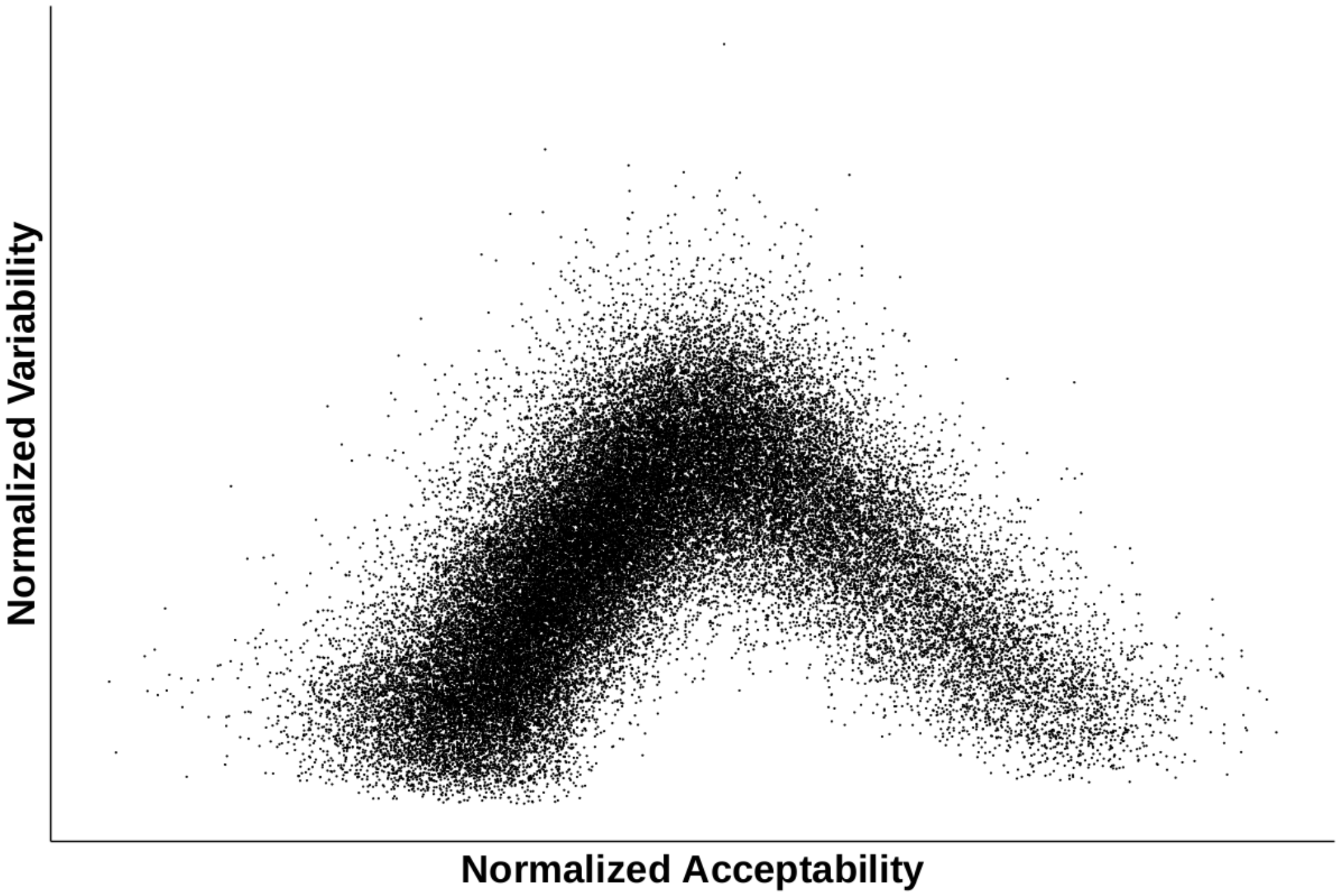

**Figure 10:** Normalized acceptability plotted against variability.

## 5.4 *Discussion*

What is the source of the models' low performance then? We believe it is likely due to a systematic bias in the kinds of information frequency distributions contain. Specifically, we posit that those aspects of a verbs' distributions that are predictable either from their abstract syntactic properties or from their meaning will not necessarily be directly encoded in their frequency distributions. That is, one will not necessarily observe all frames a particular verb is acceptable in insofar as the acceptability of that verb in that frame is predictable from its acceptability in another frame (see Grimshaw 1981; Pinker 1984; 1989; Lasnik 1989; Kako 1997; Lidz et al. 2004 for how this might work; see also Featherston 2008 on the Iceberg Phenomenon). Conversely, a verb being acceptable in a frame does not entail observing that verb in that frame, unless that acceptability is not predictable from its meaning.

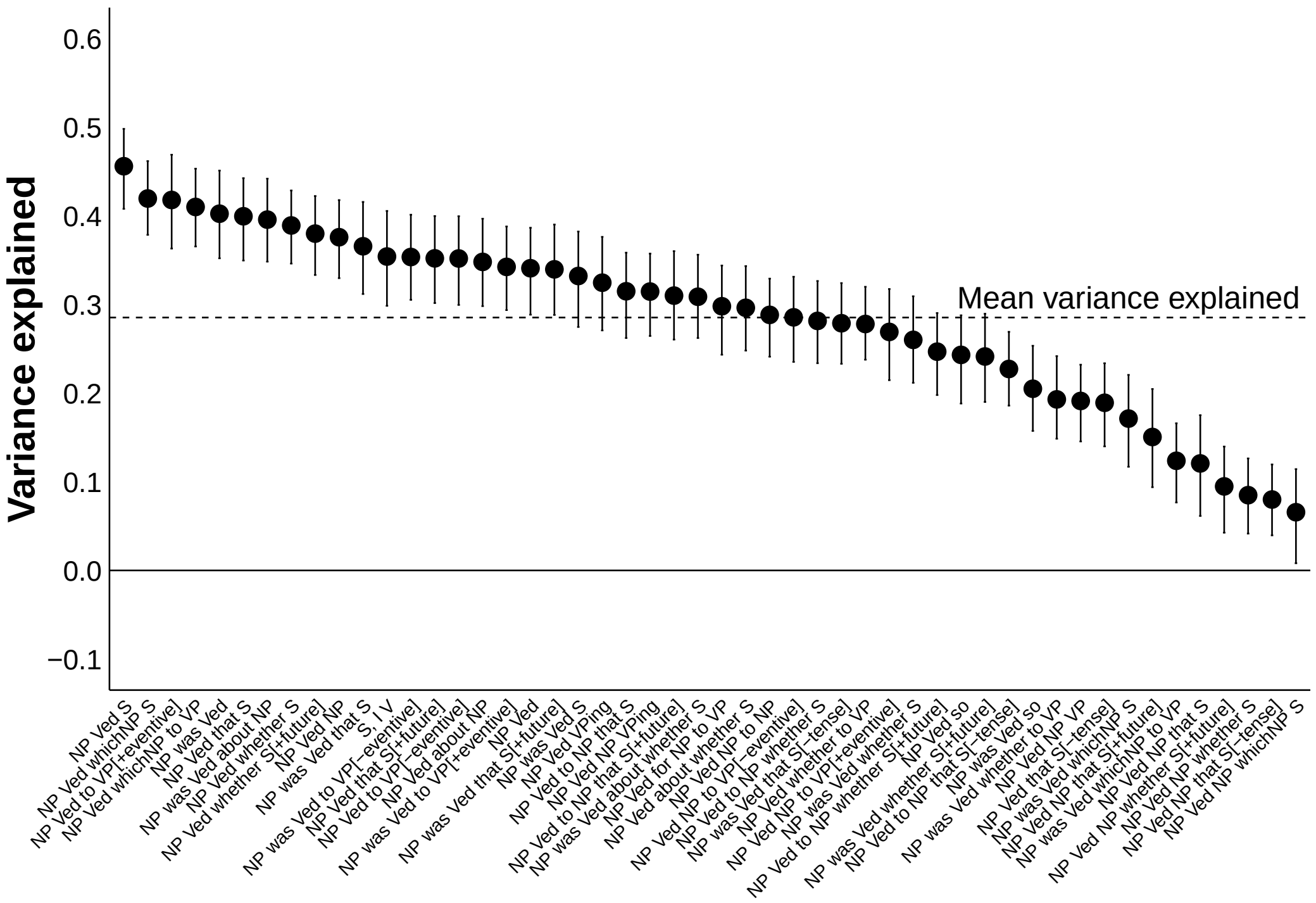

**Figure 11:** Variance explained by best-performing Beta-Negative Binomial model ($\gamma$=0.1) broken out by frame.

One piece of evidence for this comes from which frames our models perform worst on. Figure 11 plots the $R^2$ for the best-performing Beta-Negative Binomial model ($\gamma = 0.1$) broken down by frame. We see that the model systematically does more poorly in predicting verbs' acceptability in frames involving direct and indirect objects and a tensed embedded clause. This is consonant with our hypothesis insofar as verbs' acceptability in these frames is predictable from some semantic property—plausibly in this case, whether the verb is communicative or not.

Part of this hypothesis appears to fly in the face of previous work in the syntactic bootstrapping literature demonstrating that distributional cues are useful for inferring a word's meaning (Landau & Gleitman 1985; Gleitman 1990; Naigles 1990; 1996; Naigles et al. 1993; Fisher et al. 1991; Fisher 1994; Fisher et al. 1994; 2010; Lederer et al. 1995; Gillette et al. 1999; Snedeker & Gleitman 2004; Lidz et al. 2004; Gleitman et al. 2005; Papafragou et al. 2007; White 2015; White et al. 2017a; 2018a; Dudley 2017; Lewis et al. 2017). But this conflict is only apparent. A key part of our hypothesis is that acceptability is not *directly* encoded in frequency distributions. But certain components of that distribution may be observed for particular verbs, and the observation of that component may imply the acceptability of others. For instance, there is a relatively strong correlation between acceptability in an *NP Ved NP that S* frame and acceptability in an *NP Ved NP whether S* and so it is generally a safe bet that if a verb is acceptable in one it will be acceptable in another.

What this view implies is that, insofar as abstract syntactic and semantic properties reveal themselves in verb's subcategorization frame frequency distributions, it may be possible to infer those properties from regularities observable across those distributions. So far, the results are consistent only with our hypothesis H2 from Section 2, and not with its direct counterpart: we apparently *cannot* predict acceptability in selectional behavior from frequency. We consider various models of this abstraction in the next section.

# 6 Abstracting Frequency

Constructing useful abstractions of frequency distributions is a major component of much work in NLP. Abstraction techniques can take myriad forms, both probabilistic and neural. We consider four popular abstraction techniques, selected to be roughly analogous to the probabilistic and information theoretic models presented in the last section.

Our main goal in doing this is to determine the extent to which these abstractions represent acceptability directly, by which we mean that the space of abstractions and acceptability are homomorphic. As noted in the last section, the set of homomorphisms in a vector space are just the linear functions, and so as in the last section, we will attempt to predict acceptability using linear regression on the different representations we consider. We point this out because, for some of the representations we construct, it is common practice to learn nonlinear functions to a quantity of interest; and while this can be useful for understanding whether a particular abstraction implicitly contains information about a quantity, it does not tell us the extent to which that abstraction is potentially a representation *of* that quantity in an algebraic sense.

### 6.1 Models

The first model we consider is Latent Dirichlet Allocation (LDA; Blei et al. 2003), which is analogous to the Dirichlet-Categorical model presented in Section 5. This model is closely related to Alishahi & Stevenson's (2008) model of verb learning. It assumes that each verb is probabilistically associated with a set of $K$ latent syntactic and semantic properties via a conditional Categorical probability distribution $\mathbb{P}(k \mid v) = \theta_{vk}$ and that each frame is probabilistically associated with that same set of properties via a conditional Categorical probability distribution $\mathbb{P}(f \mid k) = \phi_{kf}$. The probability of seeing a verb $v$ in a frame $f$ is then modeled via these two distributions.

$$\mathbb{P}(f \mid v) = \sum_k \mathbb{P}(f \mid k)\mathbb{P}(k \mid v) = \sum_k \theta_{vk}\phi_{kf}$$

As for the Dirichlet-Categorical model from Section 5, the parameters $\boldsymbol{\theta}_v$ and $\boldsymbol{\phi}_k$ are assumed to be distributed Dirichlet.

The second model we consider is logistic factor analysis (LFA) with a negative binomial likelihood (see Zhou 2018). This model is closely related to the Poisson Factor Analysis (Zhou et al. 2012; Zhou & Carin 2015) model proposed as a model of syntactic bootstrapping by White (2015) and further developed in White et al. (2017a). This model is analogous to the Beta-Negative Binomial model presented in Section 5, using the same likelihood function but modeling $\mathbf{\Pi}$ via two matrices $\mathbf{U} \in \mathbb{R}^{N_V \times K}$ and $\mathbf{A} \in \mathbb{R}^{K \times N_F}$, where $N_V$ is the number of verbs and $N_F$ is the number of frames.

$$\pi_{vf} = \text{logit}^{-1}\left(\sum_k u_{vk}a_{kf}\right)$$

Similar to $\mathbf{\Theta}$ in LDA, one way to view $\mathbf{U}$ is as encoding verbs' abstract syntactic and/or semantic properties; and similar to $\mathbf{\Phi}$ in LDA, one way to view $\mathbf{A}$ is as encoding the syntactic properties of frame along with whatever aspect of verbal semantics project onto that frame (White & Rawlins 2016). As for the Beta-Negative Binomial model presented in Section 5, we infer $\mathbf{U}$, $\mathbf{V}$, and the rate parameters $\mathbf{r}$ of the Negative Binomial likelihood using gradient descent.

The third model we consider uses Global Vectors (GloVe; Pennington et al. 2014), which is a popular *word embedding* method in NLP. GloVe itself is not directly analogous to the PMI method from Section 5, but it is closely related (Levy & Goldberg 2014; Suzuki & Nagata 2015). In essence, it is a factor analysis of the log cooccurrence counts $c_{ij}$ for words $i$ and $j$.[20]

$$\mathbb{P}(c_{ij} \mid \mathbf{W}, \mathbf{W}', \mathbf{b}, \mathbf{b}') = \mathcal{N}(\log c_{ij}; \mathbf{w}_i \cdot \mathbf{w}'_j + b_i + b'_j)$$

As for LDA and LFA, $\mathbf{W}$ represents a relation between verbs and latent syntactic and/or semantic properties and $\mathbf{W}'$ represents the relation between these properties and frames.

We consider two versions of this GloVe-based model. The first uses pretrained GloVe to compute a neural bag of words (NBoW) representation for each sentence (Iyyer et al. 2015).[21] In NBoW, the point-wise mean of the vector for each word in a multiset is computed. In our case, this multiset is the multiset of words in each sentence of MegaAcceptability. We then predict the acceptability for that sentence from its NBoW representation.

In addition to using pretrained GloVe, we train our own GloVe embeddings on the basis of the VALEX verb-frame counts $c_{vf}$. This yields an embedding for each verb and each frame on a $K$-dimensional latent space. We consider the same settings of $K$ for this space as for the LDA and LFA models.

The final model we consider uses contextual word embeddings produced using pretrained Bidirectional Encoder Representations from Transformers (BERT; Devlin et al. 2019).[22] A full technical explication of BERT is not possible here, but in essence, BERT consists of multiple layers of interacting neural network modules known as a transformers (Vaswani et al. 2017) that are trained to predict the probability of a word given the surrounding words in the sentence as well as sentence ordering in a document. This means that BERT's representation of each word in a sentence contains some amount of information about other words in the sentence.

We use BERT to encode each sentence in MegaAcceptability and then extract the embedding of the sentence start token—i.e. the *classifier* token (`[CLS]`)—following standard practice (Devlin et al. 2019).[23] This is analo-gous to using NBoW with GloVe embeddings in that the sentence start token contains information about all words in the sentence (along with their positions) due to the way the model is trained.

[20] It diverges slightly from a standard factor analysis in downweighting the contributions of low frequency words by a factor of $f(c_{ij}) = \min\left(1, \frac{c_{ij}}{c_{\text{cutoff}}}\right)^{\alpha}$. In the pretrained models, $c_{\text{cutoff}}$ is set 100 and $\alpha$ is set to $\frac{3}{4}$. In the models we train ourselves, we set $c_{\text{cutoff}}$ to 10 and retain the same $\alpha$, since VALEX contains an order of magnitude fewer observations than the cooccurrence matrices pretrained GloVe is trained on.

[21] Pretrained GloVe is available at https://nlp.stanford.edu/projects/glove/. We specifically use the uncased, 300 dimensional vectors trained on 42 billion words of Common Crawl.

[22] Pretrained BERT models are available at https://github.com/google-research/bert. We specifically use the BERT-base-uncased models.

[23] We also experimented with extracting the embedding of the clause-embedding verb in the sentence. The results were the same.

## 6.2 Experiments

We compute MAP estimates for the parameters of LDA using the default hyperparameters in the `sklearn` package, and we compute Maximum Likelihood Estimates (MLEs) for the parameters of the logistic factor analysis model and the GloVe representations we train on VALEX. For all three models, we consider numbers of latent components $K \in \{2, 5, 10, 15, 20, 25, 30, 35, 40, 45, 50\}$. For LDA and LFA, we additionally concatenate the predicted distributions over the subcategorization frames and the best-performing normalized distributions from the Dirichlet-Categorical ($\lambda = 0$) and Beta-Negative Binomial ($\gamma = 0.1$) models, respectively.

As in Section 5, we regress the normalized acceptability judgments for each verb in each frame on each of these representations in a multivariate ridge regression—i.e. a linear regression with L2 regularization. To set the regularization parameter $\alpha \in \{0.01, 0.1, 0.2, 0.5, 1, 2, 5, 10\}$, we use a 10-fold cross-validation. To compute the generalizability of this model, we nest this cross-validation within another 10-fold cross-validation and compute the mean $R^2$ (variance explained) on the held out datasets in this outer cross-validation.

## 6.3 Results

Figure 12 plots the mean $R^2$ across the 10 cross-validation folds for each model and number of latent components. The black dashed line shows the best performing model from Section 5: Beta-Negative Binomial ($\gamma = 0.1$).

We see that BERT is far and away the best performing model with LFA a distant second. In turn, LFA ($K = 5$) outperforms the Beta-Negative Binomial model ($\gamma = 0.1$) by approximately one point—though not reliably—as well as the best-performing LDA ($K = 30$) and GloVe ($K = 15$) models by five points, reliably.

Figure 13 plots the mean $R^2$ across the 10 cross-validation folds for the BERT model, broken out by frame. We see that, in contrast to the analogous Figure 11, frames involving NP direct objects do not show systematically poor performance, suggesting that BERT may be able to capture some regularity about selection of a direct object that the models based on frame frequencies were not able to.

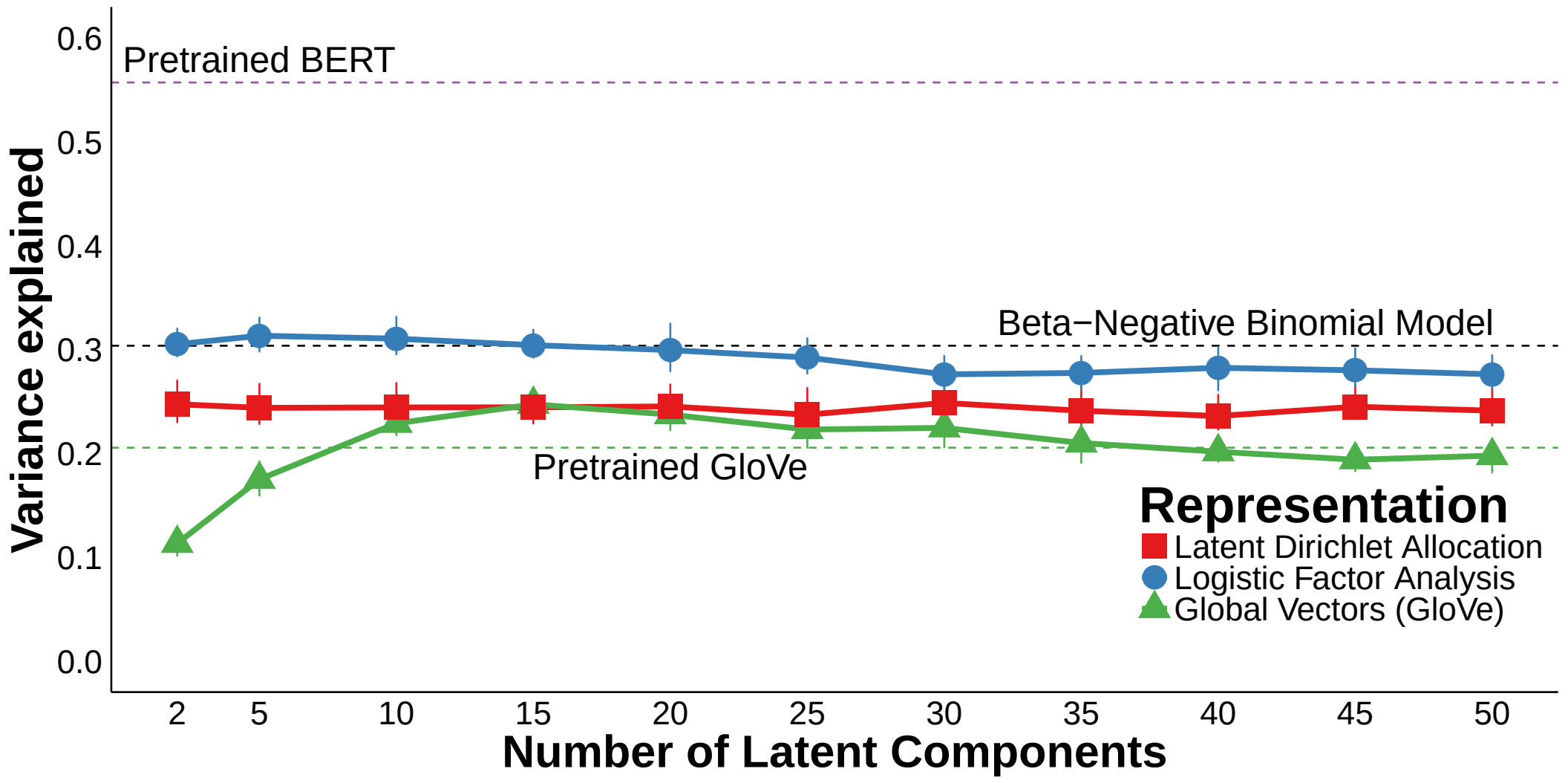

**Figure 12:** Mean variance explained in normalized acceptability judgments in 10-fold/10-fold nested cross-validation for each model and number of latent components.

## 6.4 Discussion

At a high level, these results suggest two things. First, there is some amount of information in verbs' subcategorization frame frequency distributions that is not accessible directly from those distributions themselves—even after various forms of clever normalization. Accessing that information requires some amount of abstraction of the frequencies, confirming both hypothesis H1 and H2 from Section 2.

Second, the amount of extra information that can be gleaned from the subcategorization frame frequency distributions alone is relatively small—especially compared to the gains obtained in using models, such as BERT, that additionally have access to the particular lexical items that cooccur with a verb (see Grimshaw 1994; Pinker 1994; Resnik 1996; White et al. 2017c for reasons this might be true). But not just any model of lexical cooccurrence will do, since pretrained GloVe, which does have access to such cooccurrence statistics, is one of the worst performing models of all.

These results, as of yet, do not make strong commitments as to the nature of the abstraction that confirms H2. Part of this poor performance of the shallow (non-BERT) models may be a product of the heavy constraints that we place on the space of functions that we considered from abstractions to acceptabilities. But this makes the good performance of BERT even more surprising because, though state-of-the-art performance has been demonstrated on multiple NLP tasks using its embeddings, those models invariably learn nonlinear mappings from the embeddings to the quantity of interest—as is common practice for neural methods in NLP (see Goldberg 2017 and references therein). While the results are consistent with ideas from linguistic theory and acquisition about what these abstractions might be like—BERT is typically thought to be rich enough that substantial syntactic/semantic information is present in it (Devlin et al. 2019)—it will require substantially more investigation to understand exactly how this model is predicting acceptability so well. We leave this as an open question for future work, noting that relevant (though still inconclusive) investigations exist in the NLP literature (Linzen et al. 2018; 2019).

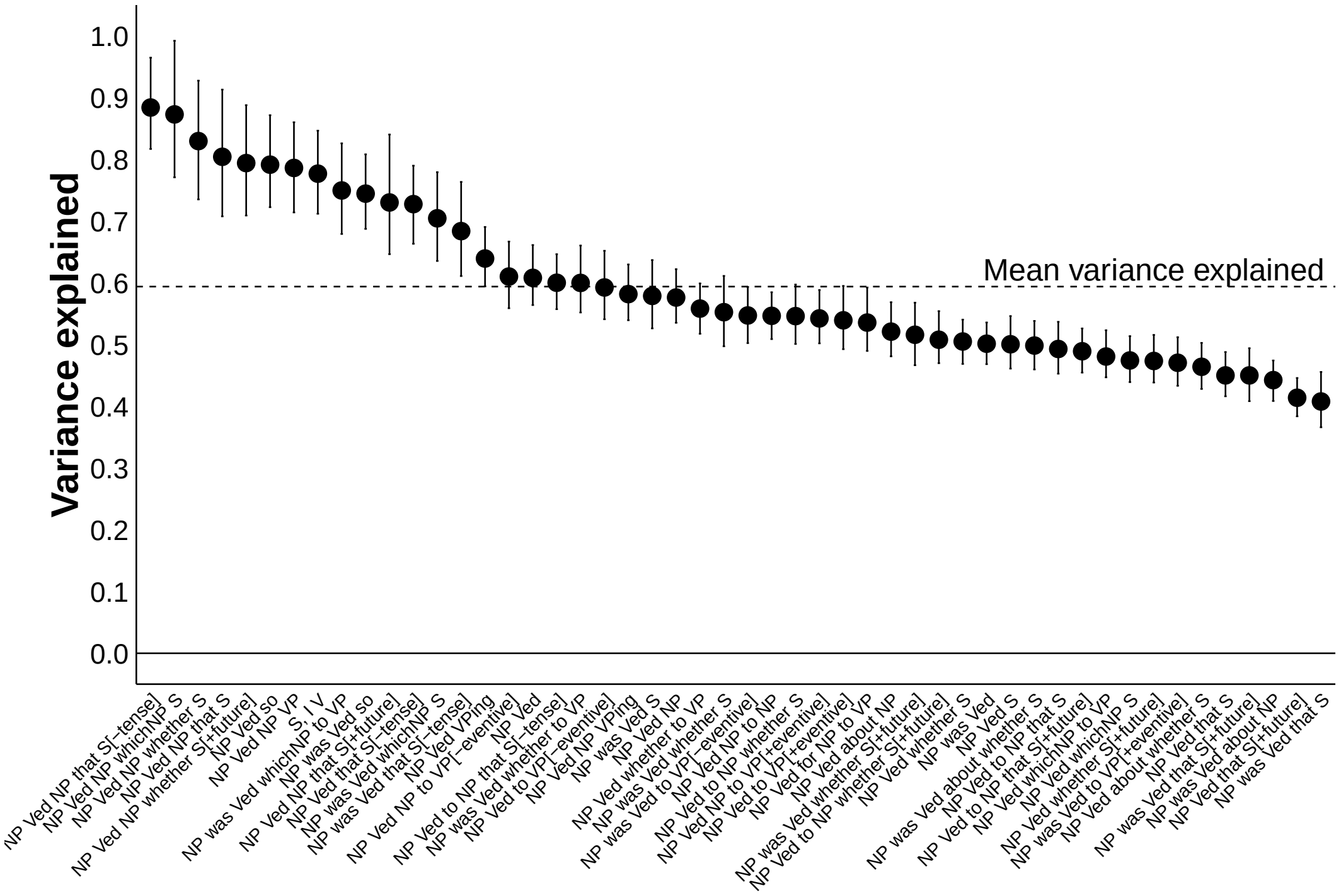

**Figure 13:** Variance explained by BERT model broken out by frame.

# 7 Conclusion

This paper addresses the question of how direct the relationship between well-formedness and linguistic experience is, focusing in particular on lexical knowledge. To do this, we developed the *bleaching method* for scaling standard acceptability judgment experiments to very large sets of verbs. After validating this method against more standard methods, we deployed it on 1,000 clause-embedding verbs in 50 syntactic frames to create the MegaAcceptability dataset, which is publicly available at megaattitude.io under the auspices of the MegaAttitude Project. Using this dataset, which we take to exhaust the set of clause-embedding verbs in English, we found that the relationship between acceptability and subcategorization frame frequency is surprisingly weak and that shallow abstractions of the data yield miniscule improvements in the prediction of acceptability. The performance of BERT suggests that deeper abstractions, however, can do surprisingly well at predicting acceptability, though we still see quite a bit of variation.

We take our results to imply that accounts of how knowledge of c-selection is acquired must posit something beyond simple smoothing or shallow factorization, as previous computational accounts have done (Alishahi & Stevenson 2008; Barak et al. 2012). One form this might take is to rely exclusively on deep, domain-general abstraction mechanisms, like BERT. Another is to enrich shallow domain-general factorization models with tunable domain-specific biases (White et al. 2017a). There is an inherent trade-off between these approaches: (i) Occam's razor implores us to posit as few inherent biases as possible; but (ii) the data hungriness of deep abstraction mechanisms makes a strictly domain-general model suspect, unless it comes with biases specific to language learning. A hybrid approach is almost certainly necessary, and we believe the MegaAcceptability dataset will prove useful in evaluating the success of such approaches.

# Supplementary files

All of the datasets collected by the authors for this paper, including the MegaAcceptability dataset, are available at megaattitude.io. All of the code necessary for replicating the analyses presented in this paper are available at megaattitude.io as well.

# Funding information

This research was funded by the following National Science Foundation grants: BCS-1748969/BCS-1749025 (*The MegaAttitude Project: Investigating selection and polysemy at the scale of the lexicon*), DDRIG BCS-1456013 (*Learning attitude verb meanings*), INSPIRE BCS-1344269 (*Gradient symbolic computation*) as well as the JHU Science of Learning Institute.

# Acknowledgements

The authors wish to thank two anonymous reviewers, Ben Van Durme, Dee Ann Reisinger, Rachel Rudinger, Charles Yang, members of the Formal and Computational Semantics Lab (FACTS.lab) at UR and the JHU Semantics Lab, and audiences at SALT 26, DGfS 2017, NELS 2017, NELS 2018, Johns Hopkins University, the University of Rochester, and Stanford University for helpful discussion of this work.

A variety of Python and R packages were used for the analyses presented in this paper, including `numpy` (Walt et al. 2011), `scipy` (Virtanen et al. 2020), `pandas` (McKinney 2011), `sklearn` (Pedregosa et al. 2011), `tensorflow` (Abadi et al. 2015), `torch` (Paszke et al. 2019), `transformers` (Wolf et al. 2019), `lme4` (Bates et al. 2015), and `turktools` (Erlewine & Kotek 2016). All plots were generated using `ggplot2` (Wickham 2016). Version information is explicitly specified in the associated analysis code available at megaattitude.io.

# Competing interests

The authors have no competing interests to declare.

# Authors' contributions

White and Rawlins collaborated on designing the materials for the MegaAcceptability dataset and writing this paper. White designed the validation experiment; implemented and conducted all experiments; and developed and implemented all models and analyses.

# References

Abadi, Martín, Ashish Agarwal, Paul Barham, Eugene Brevdo, Zhifeng Chen, Craig Citro, Greg S. Corrado, Andy Davis, Jeffrey Dean, Matthieu Devin, Sanjay Ghemawat, Ian Goodfellow, Andrew Harp, Geoffrey Irving, Michael Isard, Yangqing Jia, Rafal Jozefowicz, Lukasz Kaiser, Manjunath Kudlur, Josh Levenberg, Dandelion Mané, Rajat Monga, Sherry Moore, Derek Murray, Chris Olah, Mike Schuster, Jonathon Shlens, Benoit Steiner, Ilya Sutskever, Kunal Talwar, Paul Tucker, Vincent Vanhoucke, Vijay Vasudevan, Fernanda Viégas, Oriol Vinyals, Pete Warden, Martin Wattenberg, Martin Wicke, Yuan Yu & Xiaoqiang Zheng. 2015. TensorFlow: Large-scale machine learning on heterogeneous systems. Software available from tensorflow.org. https://www.tensorflow.org/.

Adger, David. 2003. *Core Syntax: A minimalist approach*. Oxford University Press.

Agresti, Alan. 2014. *Categorical Data Analysis*. John Wiley & Sons.

Akaike, Hirotugu. 1974. A new look at the statistical model identification. *IEEE Transactions on Automatic Control* 19(6). 716–723.

Alishahi, Afra & Suzanne Stevenson. 2008. A computational model of early argument structure acquisition. *Cognitive Science* 32(5). 789–834.

Altmann, Gerry & Yuki Kamide. 1999. Incremental interpretation at verbs: Restricting the domain of subsequent reference. *Cognition* 73(3). 247–264.

Anand, Pranav & Valentine Hacquard. 2013. Epistemics and attitudes. *Semantics and Pragmatics* 6(8). 1–59.

Anand, Pranav & Valentine Hacquard. 2014. Factivity, belief and discourse. In Luka Crnič & Uli Sauerland (eds.), *The Art and Craft of Semantics: A Festschrift for Irene Heim*, vol. 1, 69–90. Cambridge, MA: MIT Working Papers in Linguistics.

Aslin, Richard, Jenny Saffran & Elissa Newport. 1998. Computation of conditional probability statistics by 8-month-old infants. *Psychological Science* 9(4). 321–324.

Barak, Libby, Afsaneh Fazly & Suzanne Stevenson. 2012. Modeling the acquisition of mental state verbs. In *Proceedings of the 3rd Workshop on Cognitive Modeling and Computational Linguistics*. 1–10. Montreal, Canada: Association for Computational Linguistics.

Bard, Ellen Gurman, Dan Robertson & Antonella Sorace. 1996. Magnitude estimation of linguistic acceptability. *Language* 32–68.

Bates, Douglas, Martin Mächler, Ben Bolker & Steve Walker. 2015. Fitting linear mixed-effects models using lme4. *Journal of Statistical Software*

67(1). 1–48. https://doi.org/10.18637/jss.v067.i01.

Bishop, Christopher M. 2006. *Pattern recognition and machine learning*. Springer Science + Business Media.

Blei, David M., Andrew Y. Ng & Michael I. Jordan. 2003. Latent dirichlet allocation. *The Journal of Machine Learning Research* 3. 993–1022.

Bresnan, Joan. 2007. Is syntactic knowledge probabilistic? Experiments with the English dative alternation. In Sam Featherston & Wolfgang Sternefeld (eds.), *Roots: Linguistics in search of its evidential base*, vol. 96, 77–96. Walter de Gruyter.

Bresnan, Joan, Anna Cueni, Tatiana Nikitina & R Harald Baayen. 2007. Predicting the dative alternation. In Gerlof Bouma, Irene Kramer & Joost Zwarts (eds.), *Cognitive Foundations of Interpretation*, 69–94. Chicago: University of Chicago Press.

Briscoe, Ted & John Carroll. 1997. Automatic Extraction of Subcategorization from Corpora. In *Proceedings of the Fifth Conference on Applied Natural Language Processing* (ANLC '97). 356–363. Stroudsburg, PA, USA: Association for Computational Linguistics. https://doi.org/10.3115/974557.974609. https://doi.org/10.3115/974557.974609. Event-place: Washington, DC.

Callison-Burch, Chris. 2019. Crowdsourcing and human computation. http://crowdsourcing-class.org/. Accessed: 2019-11-26.

Chomsky, Noam. 1965. *Aspects of the Theory of Syntax*. Cambridge, MA: MIT Press.

Chomsky, Noam. 1973. Conditions on transformations. In S. Anderson & P. Kiparsky (eds.), *A Festschrift for Morris Halle*, 232–286. New York: Holt, Rinehart, & Winston.

Church, Kenneth W. & William A. Gale. 1995. Poisson mixtures. *Natural Language Engineering* 1(02). 163–190.

Church, Kenneth Ward & Patrick Hanks. 1990. Word association norms, mutual information, and lexicography. *Computational Linguistics* 16(1). 22–29.

Clark, Alexander, Gianluca Giorgolo & Shalom Lappin. 2013a. Statistical representation of grammaticality judgements: The limits of n-gram models. In *Proceedings of the Fourth Annual Workshop on Cognitive Modeling and Computational Linguistics*. 28–36.

Clark, Alexander, Gianluca Giorgolo & Shalom Lappin. 2013b. Towards a statistical model of grammaticality. In *Proceedings of the 35th annual conference of the cognitive science society*. 2064–2069.

Clark, Alexander & Shalom Lappin. 2011. *Linguistic Nativism and the Poverty of the Stimulus*. Wiley-Blackwell.

Conneau, Alexis, Germán Kruszewski, Guillaume Lample, Loïc Barrault & Marco Baroni. 2018. What you can cram into a single vector: Probing sentence embeddings for linguistic properties. In *Proceedings of the 56th Annual Meeting of the Association for Computational Linguistics*, vol. 1. 2126–2136. Association for Computational Linguistics. https://aclanthology.info/papers/P18-1198/p18-1198.

Devlin, Jacob, Ming-Wei Chang, Kenton Lee & Kristina Toutanova. 2019. BERT: Pre-training of deep bidirectional transformers for language understanding. In *Proceedings of the 2019 conference of the north American chapter of the association for computational linguistics: Human language technologies, volume 1 (long and short papers)*. 4171–4186. Minneapolis, Minnesota: Association for Computational Linguistics. https://doi.org/10.18653/v1/N19-1423. https://www.aclweb.org/anthology/N19-1423.

Dudley, Rachel. 2017. *The role of input in discovering presuppositions triggers: Figuring out what everybody already knew*. College Park, MD: University of Maryland dissertation.

Dunning, Ted. 1993. Accurate methods for the statistics of surprise and coincidence. *Computational Linguistics* 19(1). 61–74.

Erlewine, Michael Yoshitaka & Hadas Kotek. 2016. A streamlined approach to online linguistic surveys. *Natural Language & Linguistic Theory* 34(2). 481–495.

Featherston, Sam. 2005. Magnitude estimation and what it can do for your syntax: Some wh-constraints in German. *Lingua* 115(11). 1525–1550.

Featherston, Sam. 2007. Data in generative grammar: The stick and the carrot. *Theoretical Linguistics* 33(3). 269–318.

Featherston, Sam. 2008. Thermometer judgments as linguistic evidence. *Was ist linguistische Evidenz* 69–89.

Fillmore, Charles John. 1970. The grammar of hitting and breaking. In R.A. Jacobs & P.S. Rosenbaum (eds.), *Readings in English Transformational Grammar*, 120–133. Waltham, MA: Ginn.

Fine, Alex B. & T. Florian Jaeger. 2013. Evidence for implicit learning in syntactic comprehension. *Cognitive Science* 37(3). 578–591.

Fisher, Cynthia. 1994. Structure and meaning in the verb lexicon: Input for a syntax-aided verb learning procedure. *Language and Cognitive Processes* 9(4). 473–517.

Fisher, Cynthia, Yael Gertner, Rose M. Scott & Sylvia Yuan. 2010. Syntactic bootstrapping. *Wiley Interdisciplinary Reviews: Cognitive Science* 1(2). 143–149.

Fisher, Cynthia, Henry Gleitman & Lila R. Gleitman. 1991. On the semantic content of subcategorization frames. *Cognitive Psychology* 23(3). 331–392.

Fisher, Cynthia, D. Geoffrey Hall, Susan Rakowitz & Lila Gleitman. 1994. When it is better to receive than to give: Syntactic and conceptual constraints on vocabulary growth. *Lingua* 92. 333–375.

Frana, Ilaria. 2010. *Concealed Questions. In Search of Answers*: University of Massachusetts, Amherst dissertation.

Garnsey, Susan M., Neal J. Pearlmutter, Elizabeth Myers & Melanie A. Lotocky. 1997. The contributions of verb bias and plausibility to the comprehension of temporarily ambiguous sentences. *Journal of Memory and Language* 37(1). 58–93.

Gelman, Andrew & Jennifer Hill. 2014. *Data analysis using regression and multilevel/hierarchical models*. New York City: Cambridge University Press.

Gibson, Edward & Evelina Fedorenko. 2010. Weak quantitative standards in linguistics research. *Trends in Cognitive Science* 14(6). 233–234.

Gibson, Edward & Evelina Fedorenko. 2013. The need for quantitative methods in syntax and semantics research. *Language and Cognitive Processes* 28(1-2). 88–124.

Gillette, Jane, Henry Gleitman, Lila Gleitman & Anne Lederer. 1999. Human simulations of vocabulary learning. *Cognition* 73(2). 135–176.

Gleitman, Lila. 1990. The structural sources of verb meanings. *Language Acquisition* 1(1). 3–55.

Gleitman, Lila R., Kimberly Cassidy, Rebecca Nappa, Anna Papafragou & John C. Trueswell. 2005. Hard words. *Language Learning and Development* 1(1). 23–64.

Goldberg, Yoav. 2017. *Neural Network Methods for Natural Language Processing*, vol. 37 (Synthesis Lectures on Human Language Technologies). Morgan & Claypool Publishers.

Grimshaw, Jane. 1979. Complement selection and the lexicon. *Linguistic Inquiry* 10(2). 279–326.

Grimshaw, Jane. 1981. Form, function and the language acquisition device. In C.L. Baker & John J. McCarthy (eds.), *The Logical Problem of Language Acquisition*, 165–182. Cambridge, MA: MIT Press.

Grimshaw, Jane. 1990. *Argument Structure*. Cambridge, MA: MIT Press.

Grimshaw, Jane. 1994. Lexical reconciliation. *Lingua* 92. 411–430.

Gruber, Jeffrey Steven. 1965. *Studies in Lexical Relations*. Cambridge, MA: Massachusetts Institute of Technology dissertation.

Gulordava, Kristina, Piotr Bojanowski, Edouard Grave, Tal Linzen & Marco Baroni. 2018. Colorless green recurrent networks dream hierarchically. In *Proceedings of the 2018 conference of the north American chapter of the association for computational linguistics: Human language technologies, volume 1 (long papers)*. 1195–1205. New Orleans, Louisiana: Association for Computational Linguistics. https://doi.org/10.18653/v1/N18-1108. https://www.aclweb.org/anthology/N18-1108.

Hacquard, Valentine & Alexis Wellwood. 2012. Embedding epistemic modals in English: A corpus-based study. *Semantics and Pragmatics* 5(4). 1–29.

Hale, John. 2001. A Probabilistic Earley Parser As a Psycholinguistic Model. In *Proceedings of the Second Meeting of the North American Chapter of the Association for Computational Linguistics on Language Technologies* (NAACL '01). 1–8. Stroudsburg, PA, USA: Association for Computational Linguistics.

Hankamer, Jorge & Ivan Sag. 1976. Deep and surface anaphora. *Linguistic Inquiry* 391–428.

Heim, Irene. 1979. Concealed questions. In R. Bäuerle, U. Egli & A.v. Stechow (eds.), *Semantics from Different Points of View* (Springer Series in Language and Communication), 51–60. Springer.

Hofmeister, Philip & Ivan A. Sag. 2010. Cognitive constraints and island effects. *Language* 86(2). 366–415.

Iyyer, Mohit, Varun Manjunatha, Jordan Boyd-Graber & Hal Daumé III. 2015. Deep unordered composition rivals syntactic methods for text classification. In *Proceedings of the 53rd Annual Meeting of the Association for Computational Linguistics and the 7th International Joint Conference on Natural Language Processing (Volume 1: Long Papers)*, vol. 1. 1681–1691.

Jackendoff, Ray. 1972. *Semantic Interpretation in Generative Grammar*. Cambridge, MA: MIT Press.

Jurafsky, D. & J.H. Martin. 2009. *Speech and language processing: An introduction to natural language processing, computational linguistics, and speech recognition*. Pearson-Prentice Hall.

Kako, Edward. 1997. Subcategorization Semantics and the Naturalness of Verb-Frame Pairings. *University of Pennsylvania Working Papers in Linguistics* 4(2). 11.

Kann, Katharina, Alex Warstadt, Adina Williams & Samuel R. Bowman. 2019. Verb Argument Structure Alternations in Word and Sentence Embeddings. In *Proceedings of the Society for Computation in Linguistics (SCiL) 2019*. 287–297.

Keller, Frank. 2000. *Gradience in grammar: Experimental and computational aspects of degrees of grammaticality*: University of Edinburgh dissertation.

Kipper-Schuler, Karin. 2005. *VerbNet: A broad-coverage, comprehensive verb lexicon*: University of Pennsylvania dissertation.

Kluender, Robert & Marta Kutas. 1993. Subjacency as a processing phenomenon. *Language and cognitive processes* 8(4). 573–633.

Korhonen, Anna, Yuval Krymolowski & Ted Briscoe. 2006. A large subcategorization lexicon for natural language processing applications. In *Proceedings of LREC*, vol. 6.

Kuncoro, Adhiguna, Chris Dyer, John Hale, Dani Yogatama, Stephen Clark & Phil Blunsom. 2018. LSTMs can learn syntax-sensitive dependencies well, but modeling structure makes them better. In *Proceedings of the 56th annual meeting of the association for computational linguistics (volume 1: Long papers)*. 1426–1436. Melbourne, Australia: Association for Computational Linguistics. https://doi.org/10.18653/v1/P18-1132. https://www.aclweb.org/anthology/P18-1132.

Kush, Dave, Terje Lohndal & Jon Sprouse. 2018. Investigating variation in island effects. *Natural Language & Linguistic Theory* 36(3). 743–779.

Landau, Barbara & Lila R. Gleitman. 1985. *Language and Experience: Evidence from the Blind Child*, vol. 8. Cambridge, MA: Harvard University Press.

Lasnik, Howard. 1989. On certain substitutes for negative data. In *Learnability and Linguistic Theory*, 89–105. Springer.

Lau, Jey Han, Alexander Clark & Shalom Lappin. 2017. Grammaticality, Acceptability, and Probability: A Probabilistic View of Linguistic Knowledge. *Cognitive Science* 41(5). 1202–1241. https://doi.org/10.1111/cogs.12414. https://onlinelibrary.wiley.com/doi/abs/10.1111/cogs.12414.

Lederer, Anne, Henry Gleitman & Lila Gleitman. 1995. Verbs of a feather flock together: semantic information in the structure of maternal speech. In M. Tomasello & W.E. Merriman (eds.), *Beyond Names for Things: Young Children's Acquisition of Verbs*, 277–297. Hillsdale, NJ: Lawrence Erlbaum.

Levin, Beth. 1993. *English Verb Classes and Alternations: A preliminary investigation*. Chicago: University of Chicago Press.

Levin, Beth & Malka Rappaport Hovav. 2005. *Argument Realization*. Cambridge: Cambridge University Press.

Levy, Omer & Yoav Goldberg. 2014. Neural word embedding as implicit matrix factorization. In *Advances in Neural Information Processing Systems*. 2177–2185.

Levy, Roger. 2008. Expectation-based syntactic comprehension. *Cognition* 106(3). 1126–1177.

Lewis, Shevaun, Valentine Hacquard & Jeffrey Lidz. 2017. “Think” pragmatically: Children’s interpretation of belief reports. *Language Learning and Development* 1–23.

Lidz, Jeffrey, Henry Gleitman & Lila Gleitman. 2004. Kidz in the ’hood: Syntactic bootstrapping and the mental lexicon. In D.G. Hall & S.R. Waxman (eds.), *Weaving a Lexicon*, 603–636. Cambridge, MA: MIT Press.

Linzen, Tal, Grzegorz Chrupała & Afra Alishahi (eds.). 2018. *Proceedings of the 2018 EMNLP workshop BlackboxNLP: Analyzing and interpreting neural networks for NLP*. Brussels, Belgium: Association for Computational Linguistics. https://www.aclweb.org/anthology/W18-5400.

Linzen, Tal, Grzegorz Chrupała, Yonatan Belinkov & Dieuwke Hupkes (eds.). 2019. *Proceedings of the 2019 acl workshop blackboxnlp: Analyzing and interpreting neural networks for nlp*. Florence, Italy: Association for Computational Linguistics. https://www.aclweb.org/anthology/W19-4800.

Linzen, Tal, Emmanuel Dupoux & Yoav Goldberg. 2016. Assessing the ability of LSTMs to learn syntax-sensitive dependencies. *Transactions of the Association for Computational Linguistics* 4. 521–535. https://doi.org/10.1162/tacl_a_00115. https://www.aclweb.org/anthology/Q16-1037.

Linzen, Tal & T. Florian Jaeger. 2016. Uncertainty and expectation in sentence processing: evidence from subcategorization distributions. *Cognitive science* 40(6). 1382–1411.

Manning, Chris & Hinrich Schütze. 1999. *Foundations of Statistical Natural Language Processing*. Cambridge, MA: MIT Press.

Maye, Jessica, Janet F Werker & LouAnn Gerken. 2002. Infant sensitivity to distributional information can affect phonetic discrimination. *Cognition* 82(3). B101–B111.

McCoy, Tom, Ellie Pavlick & Tal Linzen. 2019. Right for the wrong reasons: Diagnosing syntactic heuristics in natural language inference. In *Proceedings of the 57th annual meeting of the association for computational linguistics*. 3428–3448. Florence, Italy: Association for Computational Linguistics. https://doi.org/10.18653/v1/P19-1334. https://www.aclweb.org/anthology/P19-1334.

McKinney, Wes. 2011. pandas: a foundational python library for data analysis and statistics. *Python for High Performance and Scientific Computing* 14.

McRae, Ken, Michael J. Spivey-Knowlton & Michael K. Tanenhaus. 1998. Modeling the influence of thematic fit (and other constraints) in on-line sentence comprehension. *Journal of Memory and Language* 38(3). 283–312.

Naigles, L., Henry Gleitman & Lila Gleitman. 1993. Syntactic bootstrapping and verb acquisition. In Esther Dromi (ed.), *Language and Cognition: A Developmental Perspective.* (Human Development Series), Norwood, NJ: Ablex.

Naigles, Letitia. 1990. Children use syntax to learn verb meanings. *Journal of Child Language* 17(2). 357–374.

Naigles, Letitia. 1996. The use of multiple frames in verb learning via syntactic bootstrapping. *Cognition* 58(2). 221–251.

Nathan, Lance Edward. 2006. *On the Interpretation of Concealed Questions*: Massachusetts Institute of Technology dissertation.

Papafragou, Anna, Kimberly Cassidy & Lila Gleitman. 2007. When we think about thinking: The acquisition of belief verbs. *Cognition* 105(1). 125–165.

Paszke, Adam, Sam Gross, Francisco Massa, Adam Lerer, James Bradbury, Gregory Chanan, Trevor Killeen, Zeming Lin, Natalia Gimelshein, Luca Antiga, Alban Desmaison, Andreas Kopf, Edward Yang, Zachary DeVito, Martin Raison, Alykhan Tejani, Sasank Chilamkurthy, Benoit Steiner, Lu Fang, Junjie Bai & Soumith Chintala. 2019. Pytorch: An imperative style, high-performance deep learning library. In H. Wallach, H. Larochelle, A. Beygelzimer, F. d'Alché-Buc, E. Fox & R. Garnett (eds.), *Advances in neural information processing systems 32*, 8024–8035. Curran Associates, Inc. http://papers.neurips.cc/paper/9015-pytorch-an-imperative-style-high-performance-deep-learning-library.pdf.

Pearl, Lisa & Jon Sprouse. 2013. Syntactic islands and learning biases: Combining experimental syntax and computational modeling to investigate the language acquisition problem. *Language Acquisition* 20(1). 23–68.

Pedregosa, Fabian, Gaël Varoquaux, Alexandre Gramfort, Vincent Michel, Bertrand Thirion, Olivier Grisel, Mathieu Blondel, Peter Prettenhofer, Ron Weiss, Vincent Dubourg, Jake Vanderplas, Alexandre Passos, David Cournapeau, Matthieu Brucher, Matthieu Perrot & Edouard Duchesnay. 2011. Scikit-learn: Machine Learning in Python. *Journal of Machine Learning Research* 12. 2825–2830.

Pennington, Jeffrey, Richard Socher & Christopher D. Manning. 2014. Glove: Global Vectors for Word Representation. In *Proceedings of the 2014 Conference on Empirical Methods in Natural Language Processing (EMNLP)*. 1532–1543. Doha, Qatar: Association for Computational Linguistics.

Pesetsky, David. 1982. *Paths and Categories*: Massachusetts Institute of Technology dissertation.

Pesetsky, David. 1991. Zero syntax: vol. 2: Infinitives.

Peters, Matthew, Mark Neumann, Luke Zettlemoyer & Wen-tau Yih. 2018. Dissecting contextual word embeddings: Architecture and representation. In *Proceedings of the 2018 conference on empirical methods in natural language processing*. 1499–1509. Brussels, Belgium: Association for Computational Linguistics. https://doi.org/10.18653/v1/D18-1179. https://www.aclweb.org/anthology/D18-1179.

Pinker, Steven. 1984. *Language Learnability and Language Development*. Harvard University Press.

Pinker, Steven. 1989. *Learnability and Cognition: The Acquisition of Argument Structure*. Cambridge, MA: MIT Press.

Pinker, Steven. 1994. How could a child use verb syntax to learn verb semantics? *Lingua* 92. 377–410.

Poliak, Adam, Aparajita Haldar, Rachel Rudinger, J. Edward Hu, Ellie Pavlick, Aaron Steven White & Benjamin Van Durme. 2018. Collecting Diverse Natural Language Inference Problems for Sentence Representation Evaluation. In *Proceedings of the 2018 Conference on Empirical Methods in Natural Language Processing*. Brussels, Belgium: Association for Computational Linguistics.

Rawlins, Kyle. 2013. About 'about'. *Semantics and Linguistic Theory* 23. 336–357.

Resnik, Philip. 1996. Selectional constraints: An information-theoretic model and its computational realization. *Cognition* 61(1). 127–159.

Romero, Maribel. 2005. Concealed Questions and Specificational Subjects*. *Linguistics and Philosophy* 28(6). 687–737.

Ross, John Robert. 1972. Act. In Donald Davidson & Gilbert Harman (eds.), *Semantics of Natural Language*, 70–126. Springer.

Ross, John Robert. 1973. Slifting. In Maurice Gross, Morris Halle & Marcel-Paul Schützenberger (eds.), *The Formal Analysis of Natural Languages*, 133–169. The Hague: Mouton de Gruyter.

Saffran, Jenny, Richard Aslin & Elissa Newport. 1996a. Statistical learning by 8-month-old infants. *Science* 274(5294). 1926–1928.

Saffran, Jenny, Elissa Newport & Richard Aslin. 1996b. Word segmentation: The role of distributional cues. *Journal of Memory and Language* 35(4). 606–621.

Schulte im Walde, Sabine. 2006. Experiments on the automatic induction of German semantic verb classes. *Computational Linguistics* 32(2). 159–194.

Schwarz, Gideon. 1978. Estimating the dimension of a model. *The Annals of Statistics* 6(2). 461–464.

Schütze, Carson T. & Jon Sprouse. 2014. Judgment data. In Robert J. Podesva & Devyani Sharma (eds.), *Research Methods in Linguistics*, 27–50. Cambridge University Press.

Snedeker, Jesse & Lila Gleitman. 2004. Why it is hard to label our concepts. In D. Geoffrey Hall & Sandra R. Waxman (eds.), *Weaving a Lexicon*, 257–294. Cambridge, MA: MIT Press.

Sorace, Antonella & Frank Keller. 2005. Gradience in linguistic data. *Lingua* 115(11). 1497–1524.

Spivey-Knowlton, Michael & Julie C Sedivy. 1995. Resolving attachment ambiguities with multiple constraints. *Cognition* 55(3). 227–267.

Sprouse, Jon. 2007. Continuous acceptability, categorical grammaticality, and experimental syntax. *Biolinguistics* 1. 123–134.

Sprouse, Jon. 2011. A validation of Amazon Mechanical Turk for the collection of acceptability judgments in linguistic theory. *Behavorial Research* 43. 155–167.

Sprouse, Jon & Diogo Almeida. 2013. The empirical status of data in syntax: A reply to Gibson and Fedorenko. *Language and Cognitive Processes* 28(3). 222–228.

Sprouse, Jon, Carson T. Schütze & Diogo Almeida. 2013. A comparison of informal and formal acceptability judgments using a random sample from Linguistic Inquiry 2001–2010. *Lingua* 134. 219–248.

Sprouse, Jon, Matt Wagers & Colin Phillips. 2012. A test of the relation between working-memory capacity and syntactic island effects. *Language* 88(1). 82–123.

Sprouse, Jon, Beracah Yankama, Sagar Indurkhya, Sandiway Fong & Robert C. Berwick. 2018. Colorless green ideas do sleep furiously: gradient acceptability and the nature of the grammar. *The Linguistic Review* 35(3). 575–599.

Suzuki, Jun & Masaaki Nagata. 2015. A unified learning framework of skip-grams and global vectors. In *Proceedings of the 53rd Annual Meeting of the Association for Computational Linguistics and the 7th International Joint Conference on Natural Language Processing (Volume 2: Short Papers)*, vol. 2. 186–191.

Trueswell, John C., Michael K. Tanenhaus & Christopher Kello. 1993. Verb-specific constraints in sentence processing: separating effects of lexical preference from garden-paths. *Journal of Experimental Psychology: Learning, Memory, and Cognition* 19(3). 528.

Vaswani, Ashish, Noam Shazeer, Niki Parmar, Jakob Uszkoreit, Llion Jones, Aidan N Gomez, Lukasz Kaiser & Illia Polosukhin. 2017. Attention is all you need. In *Advances in Neural Information Processing Systems*. 5998–6008.

Virtanen, Pauli, Ralf Gommers, Travis E. Oliphant, Matt Haberland, Tyler Reddy, David Cournapeau, Evgeni Burovski, Pearu Peterson, Warren Weckesser, Jonathan Bright, Stéfan J. van der Walt, Matthew Brett, Joshua Wilson, K. Jarrod Millman, Nikolay Mayorov, Andrew R. J. Nelson, Eric Jones, Robert Kern, Eric Larson, CJ Carey, İlhan Polat, Yu Feng, Eric W. Moore, Jake Vand erPlas, Denis Laxalde, Josef Perktold, Robert Cimrman, Ian Henriksen, E. A. Quintero, Charles R Harris, Anne M. Archibald, Antônio H. Ribeiro, Fabian Pedregosa, Paul van Mulbregt & SciPy 1. 0 Contributors. 2020. SciPy 1.0: Fundamental Algorithms for Scientific Computing in Python. *Nature Methods* 17. 261–272. https://doi.org/https://doi.org/10.1038/s41592-019-0686-2.

Walt, Stéfan van der, S Chris Colbert & Gael Varoquaux. 2011. The numpy array: a structure for efficient numerical computation. *Computing in Science & Engineering* 13(2). 22–30.

Wang, Alex, Amanpreet Singh, Julian Michael, Felix Hill, Omer Levy & Samuel Bowman. 2018. GLUE: A multi-task benchmark and analysis platform for natural language understanding. In *Proceedings of the 2018 EMNLP workshop BlackboxNLP: Analyzing and interpreting neural networks for NLP*. 353–355. Brussels, Belgium: Association for Computational Linguistics. https://doi.org/10.18653/v1/W18-5446. https://www.aclweb.org/anthology/W18-5446.

Warstadt, Alex, Amanpreet Singh & Samuel R. Bowman. 2019. Neural Network Acceptability Judgments. *Transactions of the Association for Computational Linguistics* 7. 625–641.

Wells, Justine B., Morten H. Christiansen, David S. Race, Daniel J. Acheson & Maryellen C. MacDonald. 2009. Experience and sentence processing: Statistical learning and relative clause comprehension. *Cognitive Psychology* 58(2). 250–271.

White, Aaron Steven. 2015. *Information and Incrementality in Syntactic Bootstrapping*. College Park, MD: University of Maryland dissertation.

White, Aaron Steven. 2019. Nothing's wrong with believing (or hoping) whether.

White, Aaron Steven, Rachel Dudley, Valentine Hacquard & Jeffrey Lidz. 2014. Discovering classes of attitude verbs using subcategorization frame distributions. In Hsin-Lun Huang, Ethan Poole & Amanda Rysling (eds.), *Proceedings of the 43rd annual meeting of the North East Linguistic*

*Society*. 249–260.

White, Aaron Steven, Valentine Hacquard & Jeffrey Lidz. 2017a. The labeling problem in syntactic bootstrapping: Main clause syntax in the acquisition of propositional attitude verbs. In Kristen Syrett & Sudha Arunachalam (eds.), *Semantics in Acquisition* (Trends in Language Acquisition Research (TiLAR)), in press. John Benjamins Publishing Company.

White, Aaron Steven, Valentine Hacquard & Jeffrey Lidz. 2018a. Semantic Information and the Syntax of Propositional Attitude Verbs. *Cognitive Science* 42(2). 416–456.

White, Aaron Steven, Pushpendre Rastogi, Kevin Duh & Benjamin Van Durme. 2017b. Inference is Everything: Recasting Semantic Resources into a Unified Evaluation Framework. In *Proceedings of the Eighth International Joint Conference on Natural Language Processing (Volume 1: Long Papers)*, vol. 1. 996–1005.

White, Aaron Steven & Kyle Rawlins. 2016. A computational model of S-selection. *Semantics and Linguistic Theory* 26. 641–663.

White, Aaron Steven & Kyle Rawlins. 2018. The role of veridicality and factivity in clause selection. In *Proceedings of the 48th Annual Meeting of the North East Linguistic Society*. to appear. Amherst, MA: GLSA Publications.

White, Aaron Steven, Philip Resnik, Valentine Hacquard & Jeffrey Lidz. 2017c. The contextual modulation of semantic information.

White, Aaron Steven, Rachel Rudinger, Kyle Rawlins & Benjamin Van Durme. 2018b. Lexicosyntactic Inference in Neural Models. In *Proceedings of the 2018 Conference on Empirical Methods in Natural Language Processing*. 4717–4724. Brussels, Belgium: Association for Computational Linguistics. http://aclweb.org/anthology/D18-1501.

Wickham, Hadley. 2016. *ggplot2: Elegant graphics for data analysis*. Springer-Verlag New York. https://ggplot2.tidyverse.org.

Wilcox, Ethan, Roger Levy, Takashi Morita & Richard Futrell. 2018. What do RNN language models learn about filler–gap dependencies? In *Proceedings of the 2018 EMNLP workshop BlackboxNLP: Analyzing and interpreting neural networks for NLP*. 211–221. Brussels, Belgium: Association for Computational Linguistics. https://doi.org/10.18653/v1/W18-5423. https://www.aclweb.org/anthology/W18-5423.

Wolf, Thomas, Lysandre Debut, Victor Sanh, Julien Chaumond, Clement Delangue, Anthony Moi, Pierric Cistac, Tim Rault, R'emi Louf, Morgan Funtowicz & Jamie Brew. 2019. Huggingface's transformers: State-of-the-art natural language processing. *ArXiv* abs/1910.03771.

Zhou, M. & L. Carin. 2015. Negative Binomial Process Count and Mixture Modeling. *IEEE Transactions on Pattern Analysis and Machine Intelligence*

37(2). 307–320. https://doi.org/10.1109/TPAMI.2013.211.

Zhou, Mingyuan. 2018. Nonparametric Bayesian Negative Binomial Factor Analysis. *Bayesian Analysis* 13(4). 1065–1093. https://doi.org/10.1214/17-BA1070. https://projecteuclid.org/euclid.ba/1510801993.

Zhou, Mingyuan, Lauren Hannah, David Dunson & Lawrence Carin. 2012. Beta-Negative Binomial Process and Poisson Factor Analysis. In *Artificial Intelligence and Statistics*. 1462–1471. http://proceedings.mlr.press/v22/zhou12c.html.

# A Materials

The mappings from abstract frames to their corresponding instantiations for both our replication of White et al. 2018a (Table 1) and the MegaAcceptability dataset (Table 2) can be found below.

| Abstract Frame | Instantiated Frame |
|---|---|
| NP Ved | Someone ____ed. |
| NP Ved NP | Someone ____ed something. |
| NP Ved NP NP | Someone ____ed someone something. |
| NP Ved NP S | Someone ____ed someone something happened. |
| NP Ved NP S notense | Someone ____ed someone something happen. |
| NP Ved NP VP | Someone ____ed someone do something. |
| NP Ved NP about NP | Someone ____ed someone about something. |
| NP Ved NP that S | Someone ____ed someone that something happened. |
| NP Ved NP that S notense | Someone ____ed someone that something happen. |
| NP Ved NP to VP | Someone ____ed someone to do something. |
| NP Ved S | Someone ____ed something happened. |
| NP Ved VPing | Someone ____ed doing something. |
| NP Ved WH S | Someone ____ed why something happened. |
| NP Ved WH to VP | Someone ____ed why to do something. |
| NP Ved about NP | Someone ____ed about something. |
| NP Ved for NP to VP | Someone ____ed for someone to do something. |
| NP Ved if S | Someone ____ed if something happened. |
| NP Ved if S notense | Someone ____ed if something happen. |
| NP Ved it that S | Someone ____ed it that something happened. |
| NP Ved it that S notense | Someone ____ed it that something happen. |
| NP Ved so | Someone ____ed so. |
| NP Ved that S | Someone ____ed that something happened. |
| NP Ved that S notense | Someone ____ed that something happen. |
| NP Ved there to VP | Someone ____ed there to be a particular thing in a particular place. |
| NP Ved to | Someone ____ed to. |
| NP Ved to NP that S | Someone ____ed to someone that something happened. |
| NP Ved to NP that S notense | Someone ____ed to someone that something happen. |
| NP Ved to VP | Someone ____ed to do something. |
| NP was Ved that S | Someone was ____ed that something happened. |
| NP was Ved that S notense | Someone was ____ed that something happen. |
| NP was Ved to VP | Someone was ____ed to do something. |
| S, I V | Something happened, I ____. |
| S, NP Ved | Something happened, someone ____ed. |
| It Ved NP WH S | It ____ed someone why something happened. |
| It Ved NP WH to VP | It ____ed someone why to do something. |
| It Ved NP that S | It ____ed someone that something happened. |
| It Ved NP that S notense | It ____ed someone that something happen. |
| It Ved NP to VP | It ____ed someone to do something. |

**Table 1:** Abstract frames and corresponding instantiated frames used in our replication of White et al. 2018a (Section 3).

| **Abstract Frame** | **Instantiated Frame** |
|---|---|
| NP Ved | Someone ____ed. |
| NP Ved NP | Someone ____ed something. |
| NP Ved NP VP | Someone ____ed someone do something. |
| NP Ved NP VPing | Someone ____ed someone doing something. |
| NP Ved NP that S | Someone ____ed someone that something happened. |
| NP Ved NP that S[+future] | Someone ____ed someone that something would happen. |
| NP Ved NP that S[-tense] | Someone ____ed someone that something happen. |
| NP Ved NP to NP | Someone ____ed something to someone. |
| NP Ved NP to VP[+eventive] | Someone ____ed someone to do something. |
| NP Ved NP to VP[-eventive] | Someone ____ed someone to have something. |
| NP Ved NP whether S | Someone ____ed someone whether something happened. |
| NP Ved NP whether S[+future] | Someone ____ed someone whether something would happen. |
| NP Ved NP whichNP S | Someone ____ed someone which thing happened. |
| NP Ved S | Someone ____ed something happened. |
| NP Ved VPing | Someone ____ed doing something. |
| NP Ved about NP | Someone ____ed about something. |
| NP Ved about whether S | Someone ____ed about whether something happened. |
| NP Ved for NP to VP | Someone ____ed for something to happen. |
| NP Ved so | Someone ____ed so. |
| NP Ved that S | Someone ____ed that something happened. |
| NP Ved that S[+future] | Someone ____ed that something would happen. |
| NP Ved that S[-tense] | Someone ____ed that something happen. |
| NP Ved to NP that S | Someone ____ed to someone that something happened. |
| NP Ved to NP that S[+future] | Someone ____ed to someone that something would happen. |
| NP Ved to NP that S[-tense] | Someone ____ed to someone that something happen. |
| NP Ved to NP whether S | Someone ____ed to someone whether something happened. |
| NP Ved to NP whether S[+future] | Someone ____ed to someone whether something would happen. |
| NP Ved to VP[+eventive] | Someone ____ed to do something. |
| NP Ved to VP[-eventive] | Someone ____ed to have something. |
| NP Ved whether S | Someone ____ed whether something happened. |
| NP Ved whether S[+future] | Someone ____ed whether something would happen. |
| NP Ved whether to VP | Someone ____ed whether to do something. |
| NP Ved whichNP S | Someone ____ed which thing happened. |
| NP Ved whichNP to VP | Someone ____ed which thing to do. |
| NP was Ved | Someone was ____ed. |
| NP was Ved S | Someone was ____ed something happened. |
| NP was Ved about NP | Someone was ____ed about something. |
| NP was Ved about whether S | Someone was ____ed about whether something happened. |
| NP was Ved so | Someone was ____ed so. |
| NP was Ved that S | Someone was ____ed that something happened. |
| NP was Ved that S[+future] | Someone was ____ed that something would happen. |
| NP was Ved that S[-tense] | Someone was ____ed that something happen. |
| NP was Ved to VP[+eventive] | Someone was ____ed to do something. |
| NP was Ved to VP[-eventive] | Someone was ____ed to have something. |
| NP was Ved whether S | Someone was ____ed whether something happened. |
| NP was Ved whether S[+future] | Someone was ____ed whether something would happen. |
| NP was Ved whether to VP | Someone was ____ed whether to do something. |
| NP was Ved whichNP S | Someone was ____ed which thing happened. |
| NP was Ved whichNP to VP | Someone was ____ed which thing to do. |
| S, I V | Something happened, I ____. |

**Table 2:** Abstract frames and corresponding instantiated frames in the MegaAcceptability dataset (Section 4; see also White & Rawlins 2016).

All of the verbs used in the MegaAcceptability dataset can be found below.

**a** abhor, absolve, accept, acclaim, accredit, acknowledge, add, address, admire, admit, admonish, adore, advertise, advise, advocate, affect, affirm, afford, affront, aggravate, aggrieve, agitate, agonize, agree, aim, alarm, alert, allege, allow, alter, amaze, amuse, analyze, anger, anguish, annotate, announce, annoy, answer, anticipate, apologize, appall, appeal, appear, appease, applaud, apply, appoint, appraise, appreciate, approach, approve, argue, arouse, arrange, articulate, ascertain, ask, assert, assess, assign, assume, assure, astonish, astound, attempt, attest, audit, authorize, awe

**b** babble, back, badger, baffle, bandy, banter, bargain, bark, be, beam, bear, befuddle, beg, begin, believe, belittle, bellow, beseech, bet, bewilder, bicker, bitch, blame, blare, blast, bleat, bless, blog, bluff, bluster, boast, boggle, bore, bother, brag, brainstorm, bribe, brief, broadcast, brood, bug, bullshit, bully, bury, buy

**c** cackle, cajole, calculate, calibrate, call, calm, care, carp, catch, categorize, cause, caution, cease, celebrate, censor, censure, certify, challenge, change, chant, characterize, charge, charm, chasten, chastise, chat, chatter, check, cheer, cherish, chide, chime, chirp, choose, chronicle, chuckle, circulate, claim, clarify, classify, clear, cloud, coach, coax, coerce, come, come around, come out, comfort, command, commence, commend, comment, commission, communicate, compel, compete, complain, compliment, comprehend, compromise, compute, conceal, concede, conceive, concern, conclude, concur, condemn, condone, confess, confide, configure, confirm, confound, confuse, congratulate, conjecture, connect, consent, consider, console, conspire, constrain, consult, contact, contemplate, contend, content, contest, continue, contract, contribute, contrive, control, convey, convince, correct, corroborate, cough, counsel, counter, cover, crack, crave, credential, cringe, criticize, croak, croon, crow, crush, cry, curse

**d** dare, daunt, daydream, daze, debate, deceive, decide, declare, decline, decree, decry, deduce, deem, defend, define, deject, delete, deliberate, delight, delineate, delude, demand, demean, demonstrate, demoralize, demystify, denounce, deny, depict, deplore, depress, deride, derive, describe, design, designate, desire, despair, despise, detail, detect, determine, detest, devastate, devise, diagnose, dictate, dig, direct, disagree, disallow, disappoint, disapprove, disbelieve, discern, discipline, disclose, disconcert, discourage, discover, discriminate, discuss, disgrace, disgruntle, disgust, dis-hearten, disillusion, dislike, dismay, dismiss, disparage, dispatch, dispel, dispirit, display, displease, disprefer, disprove, dispute, disquiet, disregard, dissatisfy, dissent, distract, distress, distrust, disturb, dither, divulge, document, doubt, draw, drawl, dread, dream, drone, dub, dupe

**e** educate, elaborate, elate, elect, electrify, elucidate, email, embarrass, embellish, embitter, embolden, emphasize, employ, enchant, encourage, end, endorse, endure, energize, enforce, engage, enjoy, enlighten, enlist, enrage, ensure, enthrall, enthuse, entice, entreat, envision, envy, establish, estimate, evaluate, evidence, examine, exasperate, excite, exclaim, excuse, exhibit, exhilarate, expect, experience, explain, exploit, explore, expose, expound, express, extrapolate

**f** fabricate, face, fake, fancy, fantasize, fascinate, fax, faze, fear, feel, feign, fess up, feud, fight, figure, figure out, file, find, find out, finish, flatter, flaunt, flip out, floor, fluster, flutter, fool, forbid, force, forecast, foresee, foretell, forget, forgive, forgo, formulate, frame, freak out, fret, frighten, frown, frustrate, fuel, fume, function, fuss

**g** gab, gall, galvanize, gamble, gasp, gather, gauge, generalize, get, giggle, gladden, glare, glean, glimpse, gloat, glorify, go, gossip, grant, grasp, gratify, grieve, grill, grimace, grin, gripe, groan, grouse, growl, grumble, grunt, guarantee, guess, guide, gurgle, gush

**h** haggle, hallucinate, handle, hanker, happen, harass, hasten, hate, hear, hearten, hedge, hesitate, highlight, hinder, hint, hire, hold, holler, hoot, hope, horrify, hound, howl, humble, humiliate, hunger, hurt, hush up, hustle

**i** identify, ignore, illuminate, illustrate, imagine, imitate, impede, impel, implore, imply, impress, incense, incite, include, indicate, indict, induce, infer, influence, inform, infuriate, initiate, inquire, inscribe, insert, insinuate, insist, inspect, inspire, instigate, instruct, insult, insure, intend, intercept, interest, interject, interpret, interrogate, interview, intimate, intimidate, intrigue, investigate, invigorate, invite, irk, irritate, isolate

**j** jabber, jade, jar, jeer, jest, joke, judge, jump, justify

**k** keep, kid, know

**l** label, lament, laud, laugh, lead, leak, learn, lecture, legislate, license, lie, like, lisp, listen, loathe, lobby, log, long, look, love, lust

**m** madden, mail, maintain, make, make out, malign, mandate, manipulate, manufacture, mark, marvel, mean, measure, meditate, meet, memorize, mention, miff, mind, minimize, misinform, misjudge, mislead, miss, mistrust, moan, mock, monitor, mope, mortify, motivate, mourn, move, mumble, murmur, muse, mutter, mystify

**n** name, narrate, nauseate, need, negotiate, nonplus, note, notice, notify

**o** object, obligate, oblige, obscure, observe, obsess, offend, offer, okay, omit, operate, oppose, ordain, order, outline, outrage, overestimate, overhear, overlook, overwhelm

**p** pain, panic, pant, pardon, pause, perceive, permit, perplex, persuade, perturb, pester, petition, petrify, phone, pick, picket, picture, piece together, pine, pinpoint, pity, placate, plan, plead, please, plot, point out, ponder, pontificate, portend, portray, posit, post, pout, praise, pray, preach, predict, prefer, prejudge, prepare, present, press, pressure, presume, presuppose, pretend, print, probe, proclaim, procrastinate, prohibit, promise, prompt, prophesy, propose, protest, prove, provoke, publicize, publish, punt, pursue, puzzle

**q** qualify, quarrel, query, question, quibble, quip, quiz, quote

**r** radio, raise, rankle, rant, rap, rationalize, rave, read, reaffirm, realize, reason, reason out, reassert, reassess, reassure, rebuke, recall, recap, reckon, recognize, recollect, recommend, reconsider, reconstruct, record, recount, recruit, rediscover, reevaluate, reexamine, regard, register, regret, regulate, reiterate, reject, relate, relax, relay, relearn, relieve, relish, remain, remark, remember, remind, reminisce, renegotiate, repeat, repent, reply, report, represent, repress, reprimand, reproach, request, require, research, resent, resolve, respect, respond, restate, result, resume, retort, retract, reveal, review, revolt, ridicule, rile, ring, rouse, rue, rule, ruminate, rush

**s** sadden, sanction, satisfy, say, scare, schedule, scheme, scoff, scold, scorn, scowl, scramble, scrawl, scream, screech, scribble, scrutinize, see, seek, seem, select, send, sense, serve, set, set about, set out, settle, shame, shape, share, shatter, shock, shoot, shout, show, showcase, shriek, shut up, sicken, sigh, sign, sign on, sign up, signal, signify, simulate, sing, sketch, skirmish, slander, smell, smile, smirk, snap, sneer, snicker, snitch, snivel, snort, snub, sob, sober, soothe, sorrow, speak, specify, speculate, spellbind, splutter, spook, spot, spout, spread, spur, sputter, squabble, squawk, squeal, stagger, stammer, stand, start, start off, startle, state, steer, stereotype, stew, stifle, stimulate, stipulate, stop, store, strain, stress, struggle, strut, study, stump, stun, stupefy, stutter, subdue, submit, suffer, suggest, sulk, summarize, summon, suppose, surmise, surprise, survey, suspect, swear, sweat, swoon

**t** tackle, take, talk, tantalize, tap, tape, taste, taunt, teach, tease, televise, tell, tempt, terrify, terrorize, test, testify, thank, theorize, think, thirst, threaten, thrill, tickle, torment, torture, tout, track, train, transmit, traumatize, trick, trigger, trouble, trust, try, turn out, tutor, tweet, type

**u** uncover, underestimate, underline, underscore, understand, undertake, unnerve, unsettle, update, uphold, upset, urge, use, utter

**v** venture, verify, vex, videotape, view, vilify, visualize, voice, volunteer, vote, vow

**w** wager, wallow, want, warn, warrant, watch, weep, weigh, welcome, wheeze, whimper, whine, whisper, whoop, will, wish, witness, wonder, worry, worship, wound, wow, write

**y** yawn, yearn, yell, yelp

# B Validation Normalization

To normalize the acceptability judgments collected in the replication experiment (Section 3), we fit an ordinal (linked logit) mixed effects model to the ratings from both datasets, with fixed effects for VERB, FRAME, and their interaction and random unconstrained cutpoints for each participant (for further background on ordinal models, see Gelman & Hill 2014; Agresti 2014). This model is implemented in `tensorflow` (Abadi et al. 2015.

This procedure is analogous to the more familiar (within linguistics) approach of $z$-scoring by participant, then taking the mean of the scores for a particular verb-frame pair. The main difference between the two methods is in how they model the way that participants make responses on the basis of some 'true' continuous acceptability. Both methods associate each par-ticipant with a different way of binning the continuous acceptability scale (usually modeled as isomorphic to the real values) to produce an ordinal response—the first bin corresponding to a 1 rating, the second corresponding to a 2 rating, etc. They differ in that $z$-scoring assumes that these bins are of equal size (for a particular participant)—the inverse of which is generally estimated via the standard deviation of the raw ordinal ratings (viewed as interval data)—whereas an ordinal model with unconstrained cutpoints (for each participant), assumes the bins can be of varying sizes.

We select the particular normalization method we use on the basis of empirical findings presented in White et al. 2018a (the paper whose data we validate against in Section 3). White et al. compare the fit to their data of six different possible ordinal models, varying in 3 respects: (i) whether the bins corresponding to each rating are of constant size or vary in size; (ii) whether the bins are centered around 0 for all participants or each participant has a different center (*additive* participant effects); and (iii) whether the size of the bins stays constant across participants or can be expanded or contracted depending on the participant (*multiplicative* participant effects). They point out that $z$-scoring corresponds to the model wherein the bins are of constant size but where there are both additive and multiplicative participant effects.

They fit each of these models with fixed effects for VERB, FRAME, and their interaction—effectively, each pairing of a verb $v$ and a frame $f$ is associated with some continuous acceptability value $a_{vf} = \beta_v + \beta_f + \beta_{vf}$, which is jointly optimized with parameters representing the bins.[24] They find that, even after penalizing for model complexity using both the Akaike Information Criterion (Akaike 1974: AIC;) and the Bayesian Information Criterion (BIC; Schwarz 1978), the model with varying bin sizes and additive and multiplicative participant effects fits the data substantially better than any other model, including the one corresponding to the assumptions of $z$-scoring (constant bin sizes and additive and multiplicative participant effects). We thus use a normalization method that assumes varying bin sizes.

We parameterize this method by assuming that each pairing of a verb $v$ and a frame $f$ is associated with some true real-valued acceptability $a_{vf}$ (as described above) and that each participants $p$ is associated with a way of binning these real-valued acceptability judgments, where each bin corresponds to a particular scale rating. These bins are defined by cutpoints $\mathbf{c}_p$ for each participants $p$, where the bin corresponding to the worst rating—in our case, 1—is to the interval $(-\infty, c_{p1}]$ and the bin corresponding to the best ratings—in our case, 7—is the interval $(c_{p6}, \infty)$. For all other ratings $i$, the corresponding bin for participant $p$ is $(c_{p(i-1)}, c_{pi}]$. Alternatively, we say that $c_{p0} = -\infty$ and $c_{p7} = \infty$ for all participants $p$.

[24] Steps must be taken to ensure identifiability, but how this is done is not important for current purposes.

Similar to a binary logistic regression, which one can think of as having just two bins defined by a single cut point, we define the probability of a particular participant $p$ giving a response $r_{pvf}$ to verb $v$ and frame $f$ (assuming true acceptability $a_{vf}$) based on these cutpoints. First, we define the cumulative density function.

$$\mathbb{P}(r_{pvf} \leq i) = \text{logit}^{-1}\left(c_{pi} - a_{vf}\right)$$

Then, from the cumulative density function, we can reconstruct the probability for each response $i$.

$$\mathbb{P}(r_{pvf} = i) = \mathbb{P}(r_{pvf} \leq i) - \mathbb{P}(r \leq (i-1))$$

From this, the (log-)likelihood of the data immediately follows. This likelihood is the measure we use as a measure of variability in the main text, since the lower this likelihood is for a particular verb-frame pair, the less able the model is to 'explain' the participants' responses using a single value $a_{vf}$, even after adjusting for differences in how the participant bins the scale.

We estimate the true acceptabilities **A** for all verb-frame pairs and the cutpoints for all participants **C** by using gradient descent to maximize the sum of the likelihood of the data, an Exponential prior on the distance between the cutpoints (thereby making this a mixed effects model), and a small smoothing term, under the constraint that the mean of the third cutpoint is locked to zero, thus making the parameters identifiable. All analyses use the resulting acceptabilities **A**.

A reader may still wonder if there are empirical consequences to this choice of normalization method in contrast to $z$-scoring, even if this normalization is better theoretically and empirically motivated. In Appendix C we briefly explore this further, and show that using $z$-scoring produces scores that are highly correlated with the ordinal model-based method in the data at issue here.

# C MegaAcceptability Normalization

As for our replication of White et al.'s dataset, to measure interannotator agreement, we compute the Spearman rank correlation between the responses for each pair of participants that did the same list. This yields a mean correlation of 0.416 (95% CI: [0.413, 0.419]), which is more than 10 points lower than the agreement obtained in the replication.

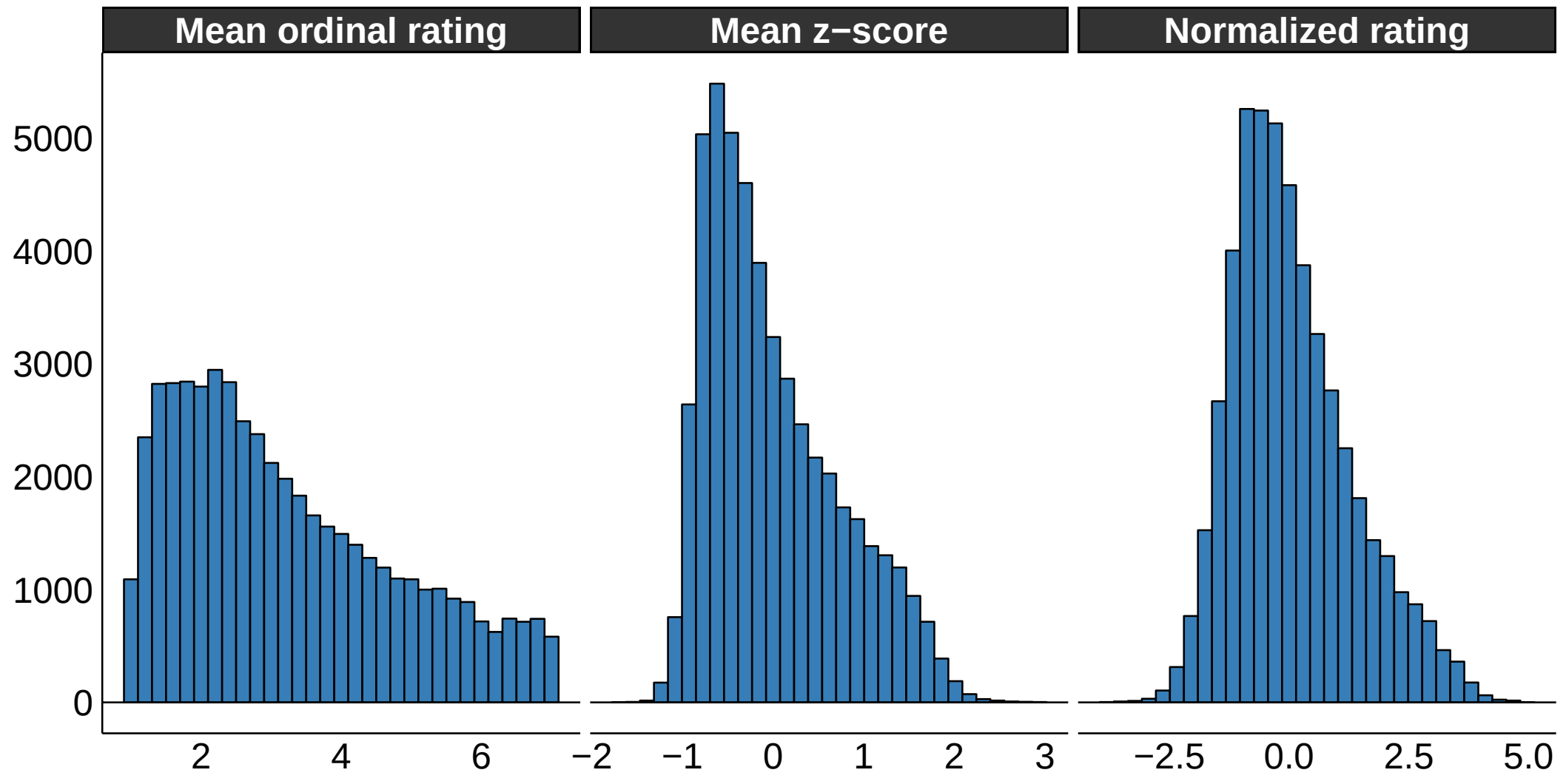

**Figure 14:** Marginal distribution across all verb-frame pairs of different acceptability scores.

Part of the reason for this is likely that White et al.'s—and consequently, our replication—contained mostly high frequency verbs, whereas the MegaAttitude dataset contains many low frequency verbs that participants are likely less certain about. Another source of this low agreement is likely a higher rate of poor participants in these data. This is evidenced by the fact that the agreement scores have nontrivial left skew, with a median correlation of 0.455 (95% CI: [0.451, 0.458]).

To mitigate the effect of poor participants, we downweight the influence of those participants' responses in constructing the normalized acceptability for each verb-frame pair. Our approach amounts to using the ordinal model-based normalization described in Section 3, but weighting the likelihood of the ordinal model by participant quality scores on [0, 1] derived from pairwise agreement between participants.[25]

One simple way of deriving such a score would be to take the mean interannotator agreement for all pairs an participant occurs in and then normalize those means to lie on [0, 1]. This simple approach is problematic, however, since most participants only rate one list and so, if a good participant rates a list rated by mostly bad participants, that participant will be assigned a low quality score.

[25] This procedure differs from the procedure used by White & Rawlins (2016) for the same dataset in that they filter participants with agreement under a particular threshold. Our approach can be seen as a soft version of their thresholding approach, wherein the influence of participants' responses drops off smoothly as a function of their overall agreement with other participants.

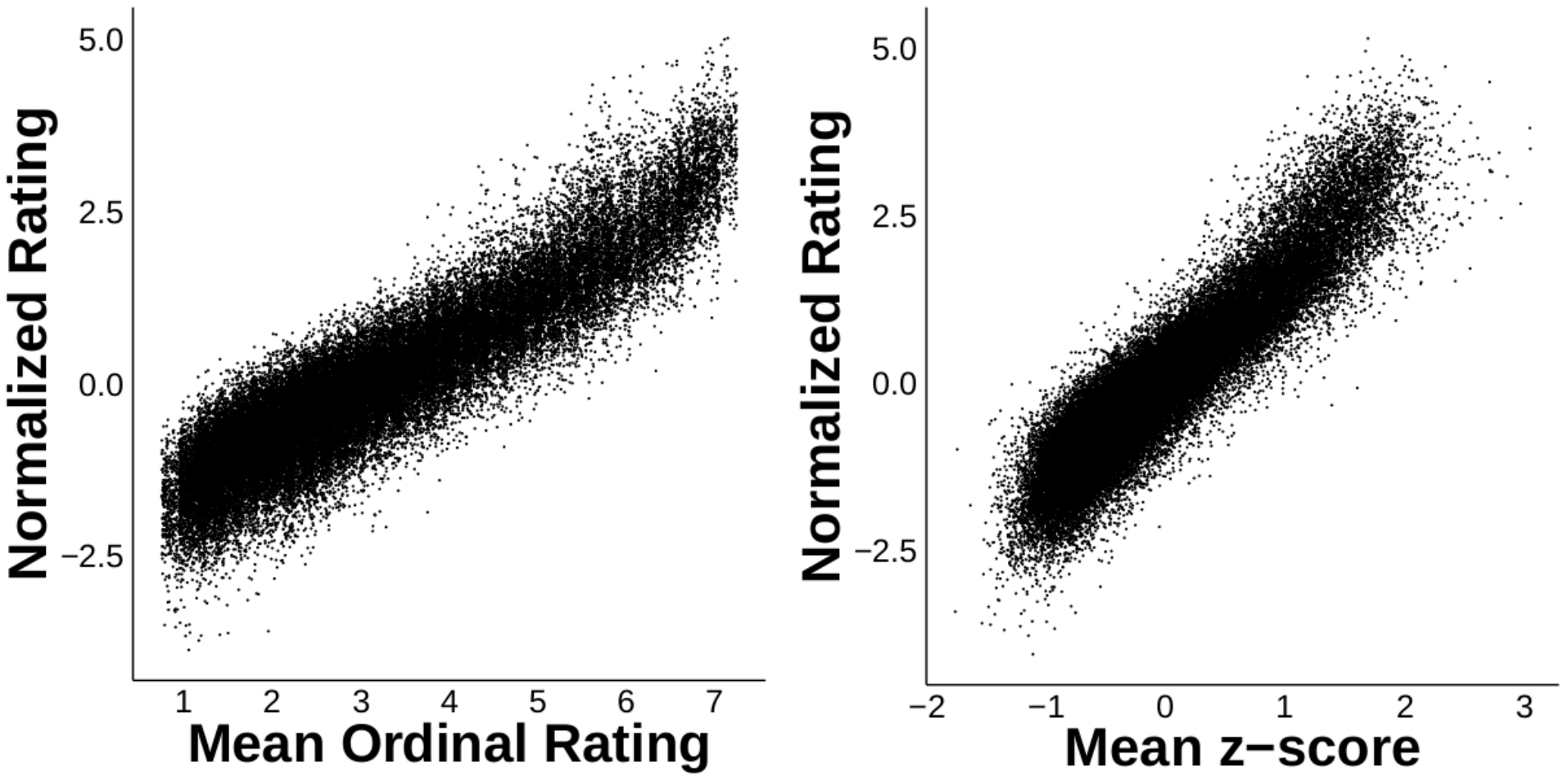

**Figure 15:** Relationship between mean ordinal responses (viewed as interval data) and normalized ratings produced by ordinal mixed model for particular verb-frame pairs (left) and relationship between mean of responses $z$-scored by participant and normalized ratings produced by ordinal mixed model for particular verb-frame pairs (right). Each point corresponds to a verb-frame pair.

To address this issue, we derive a participant quality score by first fitting a linear mixed effects model with random intercepts for participant and list to the Spearman rank correlations—using `lme4` (Bates et al. 2015)—then extracting the Best Linear Unbiased Predictors for the participant intercepts. We then $z$-score these scores and squash them to [0, 1] using the normal cumulative distribution function. This participant quality score is thus high when an participant tends to show high agreement with other participants, adjusting for the effect of the particular list.

We combine these log-likelihoods into single variability score by computing their mean, weighted by the participant quality score of the participant who provided the rating.

Figure 14 shows the marginal distribution of ratings using the above method as well as two other common methods: (i) taking the mean of the ordinal responses (viewed as interval data) for each verb-frame pair (*mean* *ordinal rating*); and (ii) taking the mean of the ratings $z$-scored by participant for each verb-frame pair.

Figure 15 plots the corresponding joint distributions—i.e. the relationship between the resulting normalized value for each verb-frame pair and the mean of the ordinal responses for that pair (left) as well as the mean of the responses $z$-scored by participant (right). The Pearson correlation between the normalized value for each verb-frame pair and the mean of the ordinal responses (viewed as interval data) for that pair is 0.92, and the correlation between the normalized value for each verb-frame pair and the mean of the responses $z$-scored by participant is 0.95.

# D Method for Adding Verbs

Seven verbs—*manage*, *fail*, *neglect*, *refuse*, *help*, *opt*, *deserve*—were unintentionally excluded from our large-scale experiment due to a coding error. We do not include these verbs in the analyses presented in the body of the paper because it is nontrivial, within the method described above, to build lists that include them without reconducting a large portion of the study.

Because we would like to have data about these verbs for future work, we instead evaluate an alternative method for adding missing verbs to our dataset. In this method, we test a single verb in all of the frames of interest within the same list.

To evaluate how this method compares to to a method wherein verbs are intermixed, we constructed a list for each of the 30 pilot verbs from Section 3 paired with each of the 50 frames from the MegaAcceptability data (Section 4). We find that the average pairwise agreement by list is actually higher in this experiment than in our original replication, with a median Spearman rank correlation of 0.65 (95% CI = [0.63 , 0.67]). This higher agreement is due to a few annotators who did many lists showing high agreement with each other, since when we fit the linear mixed effects model described in Appendix C to these correlations, we find an expected correlation of 0.54, which is very close to the correlation found in our validation experiments (Section 3).

To compare the agreement between the normalized ratings from the MegaAcceptability dataset to those from this one-verb-per-list dataset, we applied the normalization used for the MegaAcceptability dataset (Appendix C) to these data and then computed the correlation by verb. Figure 16 shows this agreement which is extremely high across all verbs.

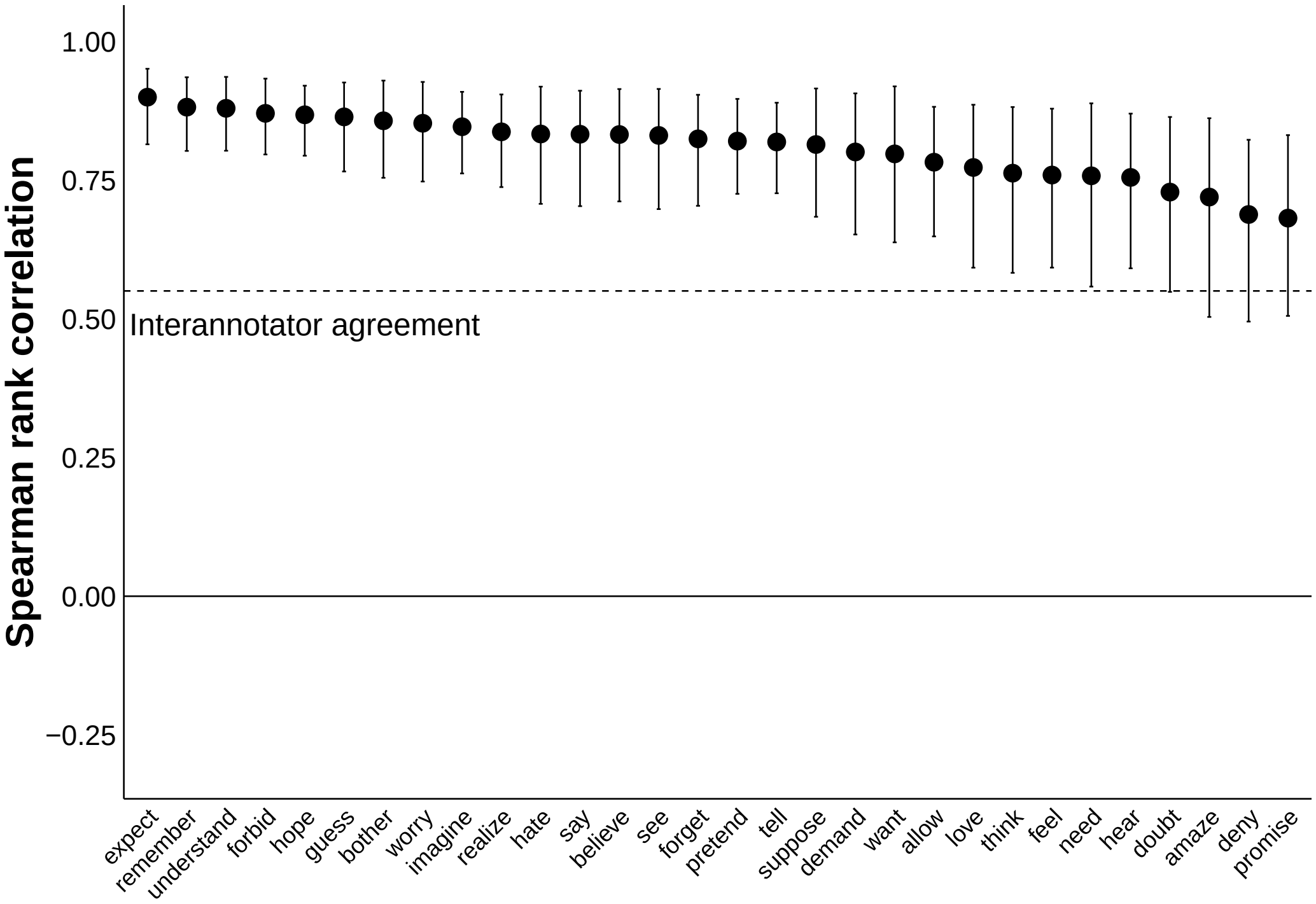

**Figure 16:** Correlation by verb between mean normalized verb-frame acceptability in MegaAcceptability and one-verb-per-list dataset. The dashed line shows mean interannotator agreement.

We take this as an indicator that testing one verb per list—at least in this set of frames—produces results that are just as valid as intermixing verbs. We thus tested the seven verbs above using this method. The resulting dataset is available on megaattitude.io.